\documentclass[final,5p,times,twocolumn,authoryear]{elsarticle}
\usepackage{url}
\usepackage{hyperref} 
\hypersetup{colorlinks = true, 
	    linkcolor = black, 
	    urlcolor = blue,
            citecolor = blue} 

\usepackage{amssymb}
\usepackage{amsmath}
\usepackage{tabularray}
\usepackage{subcaption}
\usepackage{flushend}
\usepackage{amsthm}
\usepackage[switch]{lineno}

\newtheorem{remark}{Remark}

\let\oldequation\equation
\let\oldendequation\endequation

\renewenvironment{equation}
  {\linenomathNonumbers\oldequation}
  {\oldendequation\endlinenomath}

\usepackage{lineno}

\usepackage{caption}
\journal{ISPRS Journal of Photogrammetry and Remote Sensing}  

\begin{document}

\begin{frontmatter}



\title{Reliable-loc: Robust sequential LiDAR global localization in large-scale street scenes based on verifiable cues} 


\author[label1]{Xianghong Zou}\ead{ericxhzou@whu.edu.cn}
\author[label2]{Jianping Li\corref{cor1}}\ead{jianping.li@ntu.edu.sg}
\author[label3]{Weitong Wu}\ead{weitongwu@hhu.edu.cn}
\author[label1]{Fuxun Liang}\ead{liangfuxun@whu.edu.cn}
\author[label1]{Bisheng Yang\corref{cor1}}\ead{bshyang@whu.edu.cn}
\author[label1,label4]{Zhen Dong}\ead{dongzhenwhu@whu.edu.cn}

\cortext[cor1]{Corresponding author}

\affiliation[label1]{organization={State Key Laboratory of Information Engineering in Surveying, Mapping and Remote Sensing, Wuhan University},
            city={Wuhan},
            postcode={430079}, 
            country={China}}
\affiliation[label2]{organization={School of Electrical and Electronic Engineering, Nanyang Technological University},
            city={50 Nanyang Avenue},
            postcode={639798}, 
            country={Singapore}}
\affiliation[label3]{organization={School of Earth Sciences and Engineering, Hohai University},
            city={Nanjing},
            postcode={211100}, 
            country={China}}
\affiliation[label4]{organization={Hubei Luojia Laboratory},
            city={Wuhan},
            postcode={430079}, 
            country={China}}

\begin{abstract}
Wearable laser scanning (WLS) system has the advantages of flexibility and portability. It can be used for determining the user's path within a prior map, which is a huge demand for applications in pedestrian navigation, collaborative mapping, augmented reality, and emergency rescue. However, existing LiDAR-based global localization methods suffer from insufficient robustness, especially in complex large-scale outdoor scenes with insufficient features and incomplete coverage of the prior map. To address such challenges, we propose LiDAR-based reliable global localization (Reliable-loc) exploiting the verifiable cues in the sequential LiDAR data. First, we propose a Monte Carlo Localization (MCL) based on spatially verifiable cues, utilizing the rich information embedded in local features to adjust the particles' weights hence avoiding the particles converging to erroneous regions. Second, we propose a localization status monitoring mechanism guided by the sequential pose uncertainties and adaptively switching the localization mode using the temporal verifiable cues to avoid the crash of the localization system. To validate the proposed Reliable-loc, comprehensive experiments have been conducted on a large-scale heterogeneous point cloud dataset consisting of high-precision vehicle-mounted mobile laser scanning (MLS) point clouds and helmet-mounted WLS point clouds, which cover various street scenes with a length of over 30 km. The experimental results indicate that Reliable-loc exhibits high robustness, accuracy, and efficiency in large-scale, complex street scenes, with a position accuracy of ±2.91 m, yaw accuracy of ±3.74 degrees, and achieves real-time performance. For the code and detailed experimental results, please refer to \url{https://github.com/zouxianghong/Reliable-loc}.
\end{abstract}



\begin{keyword}
LiDAR \sep Global localization \sep Monte Carlo Localization \sep Spatial verification \sep Pose uncertainty


\end{keyword}

\end{frontmatter}



\section{Introduction}\label{sec_introduction}\
Wearable laser scanning (WLS) systems have the advantages of flexibility and portability by integrating LiDAR (Light Detection And Ranging), IMU (inertial measurement unit), and other sensors in a portable device \citep{WHU_Helmet, li2024hcto}. It can be used for finding the user's path, i.e. global localization, which is a huge demand for pedestrian navigation \citep{Baglietto_Human}, collaborative localization and mapping \citep{yuan2017cooperative, WeCo_SLAM}, augmented reality \citep{Chi_Rebar}, counter-terrorism and emergency rescue \citep{bin2017evaluation}. As LiDAR is not sensitive to changes in lighting, works in different weather conditions, and has high measurement accuracy \citep{Radosevic_Laser_scanning, Sharif_Laser_privacy}, LiDAR-based global localization is practical and can work in GNSS-denied complex environments such as urban canyons, indoors, and underground \citep{GM-Livox}.

LiDAR-based global localization has been long studied. According to whether sequential LiDAR data is used, LiDAR-based global localization can be divided into two categories: single-shot and sequential global localization. The former usually achieves localization based on place recognition using a single LiDAR frame and is not suitable for large-scale repetitive scenes \citep{Dh3d, Minkloc3d, Lcdnet, PatchAugNet}. The latter usually realizes localization by utilizing place recognition with techniques like MCL and sequential matching, which is more applicable to industry applications in large-scale outdoor scenes \citep{Seqlpd, LocNet, overlap_loc, SeqOT}. Hence, this paper focuses on the sequential global localization for the wearable device.

In real applications, the WLS-based localization system needs to be operated in various scenes, posing many challenges to the reliability of the global localization algorithm. First, the localization system faces challenges from feature-insufficient environments. Take the typical urban street scene in southern China as an example: the road is wide and full of viaducts, and many road sections are only flanked by street trees and flat facades, lacking sufficient features \citep{WHU_Helmet, han2024whu}. Second, the prior map may not fully cover the area, and blank data holes exist. For example, the prior point cloud map is usually collected along roads by the vehicle-mounted MLS systems and can not fully cover roadside areas due to occlusion and accessibility \citep{serna2013urban_accessibility, mi2021automated}. In recent years, some LiDAR-based sequential global localization methods have integrated place recognition techniques into MCL and are applicable to large-scale outdoor scenes \citep{Global_LiDAR_Loc_Survey}. However, existing methods still have limitations facing the above challenges. On the one hand, most of them solely rely on the global feature (the aggregation of local features) extracted from local maps to construct the observation model in MCL. In feature-insufficient scenes where global features lack descriptive adequacy, particles in MCL are prone to converging towards erroneous regions, ultimately resulting in the localization system crashing. While many place recognition methods extract local features simultaneously to aggregate the global feature, the rich information embedded in local features is ignored by the downstream MCL task. On the other hand, existing methods only rely on point cloud registration for localization once MCL converges, thus leading to localization failure due to continuous unreliable pose estimation in the scenes with insufficient features and incomplete map coverage.

To tackle the challenges of feature insufficiency and incomplete coverage of prior maps, we propose Reliable-loc, a reliable sequential global localization method based on verifiable cues from both spatial and temporal aspects. Reliable-loc enhances MCL by exploiting the rich information embedded in local features from place recognition (\textbf{the spatial cues}) and improving localization robustness by monitoring the sequential localization status (\textbf{the temporal cues}). The main contributions of this paper are as follows:
\begin{enumerate}[1)]
    \item A novel MCL incorporating spatially verifiable cues is proposed to adjust particle weights using the rich information embedded in local features. It improves localization robustness in feature-insufficient scenes by avoiding particles converging to erroneous regions.
    \item A localization status monitoring mechanism guided by the temporal sequential pose uncertainties is proposed thus the localization mode can be adaptively switched  according to the temporal verifiable cues. It improves localization robustness in scenes with insufficient features and incomplete map coverage by exploiting the exploratory capability of particles in MCL.
\end{enumerate}

The remainder of the paper is organized as follows. Section \ref{sec_related_work} reviews the related works on LiDAR-based sequential global localization. Section \ref{sec_preliminary} presents the important preliminary. Section \ref{sec_method} elaborates on the proposed approach. Section \ref{sec_experiment} presents the datasets and quantitative evaluation. Section \ref{sec_analysis_discussion} presents the ablation studies and discusses the deficiencies and future work. The conclusion is outlined in Section \ref{sec_conclusion}.

\section{Related Work}\label{sec_related_work}
\subsection{LiDAR-based sequential place recognition}\label{subsec_seq_pr}

In recent years, various LIDAR-based sequential place recognition methods have been proposed, drawing inspiration from related works in the field of vision, such as FAB-MAP \citep{FAB_MAP} and SeqSLAM \citep{SeqSLAM}. SeqLPD \citep{Seqlpd} extracts features from point clouds via LPDNet \citep{Lpd_net}, clusters global features in the reference map using Kmeans++ \citep{k_means_plus} to obtain keyframes, and designs a coarse-to-fine sequence matching strategy inspired by SeqSLAM \citep{SeqSLAM}, enhancing the robustness of place recognition. SeqSphereVLAD \citep{Seqspherevlad_v1, Seqspherevlad_v2} first extracts features from point clouds using a spherical convolution-based network, and then achieves reliable loop closure detection by performing coarse-to-fine sequence matching based on particle filtering. SeqOT \citep{SeqOT} extracts features from a single frame point cloud based on OverlapNet \citep{OverlapNet} and then utilizes Transformer \citep{Transformer} and NetVLAD \citep{NetVLAD} to extract features from sequential point clouds. P-GAT \citep{P_GAT} proposes a pose graph attention network, taking inspiration from SuperGlue \citep{Superglue}, effectively improving place recognition by encoding temporal and spatial information within and across submap sequences. Such methods use sequence matching or feature interaction to enhance single-shot place recognition, providing the most likely place on the map rather than an accurate pose.

\subsection{LiDAR-based sequential-metric global localization}\label{subsec_seq_gl}
LiDAR-based sequential-metric global localization typically integrates place recognition into MCL and can provide an accurate pose based on metric maps \citep{Global_LiDAR_Loc_Survey}. LocNet \citep{LocNet} initiates by extracting global features from constructed histograms via deep networks, subsequently integrating them into MCL for 2D pose estimation. After MCL is converged, it uses ICP \citep{ICP} to estimate the 3D pose. However, poor global feature descriptive capability and reliance only on ICP for localization after MCL is converged impacts localization robustness. Overlap-loc \citep{overlap_loc} leverages OverlapNet \citep{OverlapNet} to regress point cloud overlap and utilizes the overlap as an observation for MCL. Despite its simplicity and efficiency, solely relying on MCL cannot achieve accurate localization in large-scale, complex scenes. Localization \citep{Localising_Faster} Faster projects point clouds onto BEV (Bird's Eye View) images, directly regressing 6-DOF poses via networks. It then employs Gaussian process regression to determine the mean and covariance of the predicted pose, facilitating particle initialization and resampling in MCL. However, direct regression of pose and associated mean and covariance is unreliable. Hybrid localization \citep{Akai_Hybrid} achieves end-to-end 2D localization by modeling the posterior distribution through a CNN (Convolutional Neural Networks) network to guide the particle sampling in MCL. \citet{Li_Robust_Loc} argues that rods in the scene exhibit stability and can be used to generate feature maps for MCL. However, this assumption falters in repetitive, symmetric scenes. DSOM \citep{DSOM} uses deep networks to directly regress the posterior distribution of the system's pose to guide particle sampling and proposes a trusty mechanism to adaptively adjust the particles' weights. SeqPolar \citep{Seqpolar} first extracts features from polar images projected from point clouds using a sophisticated algorithm, then retrieves the vehicle's approximate location using HMM2 (second-order hidden Markov model), and finally achieves accurate localization through point cloud registration. However, it is computationally inefficient and suitable only for small scenes.

To sum up, LiDAR-based sequential-metric global localization methods offer efficient localization devoid of initial poses, delivering accuracy up to a certain confidence threshold. Nevertheless, enhancing localization resilience in expansive, intricate environments remains a significant challenge necessitating continued research.

\subsection{Localization failure detection}\label{subsec_loc_failure}

Localization failure detection plays a crucial role in identifying instances where the localization system experiences a crash, thereby significantly bolstering the reliability of localization systems. However, there exists a noticeable gap in research focusing on failure detection for LiDAR-based localization.  \citet{Self_adaptive_MCL} performs an adaptive function to avoid localization failure by measuring the probability of particles. \citet{Fujii_Detection} proposes a failure detection method for MCL based on logistic regression. \citet{Alsayed_Failure} detects localization failures using extracted lines and curves. \citet{Almqvist_Learning} proposes a method based on NDT (Normal Distribution Transform) for detecting misaligned point clouds and predicting the degree of misalignment using machine learning. \citet{Yin_A_failure_detection} proposes a statistical learning-based approach for localization failure detection by defining the problem as a binary classification task. \citet{Kirsch_Predicting} computes distance measures consisting of overlap and feature distance and employs SVM (support vector machine) to classify whether point clouds are aligned or not. Of these methods, those that detect localization failures by assessing particle probabilities in MCL tend to be computationally intensive. On the other hand, methods that identify failures by assessing the alignment of point clouds via metrics like overlap, RMS (root mean square), Hessian matrix, etc., are prone to being sensitive to point density.

\section{Problem Definition and Preliminary}\label{sec_preliminary}
Let the prior point cloud map be $M=\left\{M_{i}\right\}$, where $M$ is a set consisting of point cloud submaps in the map. The sequence of point cloud submaps $Z_{1: t}$ are acquired by the WLS system up to time $t$, and its motions $U_{1: t}$ are up to time $t$. The estimation of the system's poses $\hat{X}_{1: t}$, in the global localization task, can be formulated in terms of Bayes' law as follows:
\begin{equation}
    \hat{X}_{1: t}=\operatorname{argmax} p\left(\left\{Z_{1: t}, U_{1: t}\right\} \mid X_{1: t}, M\right).
\end{equation}
Then, the estimated pose $\hat{X}_{t}$ of the system at the time $t$ can be formulated according to the Markov process:
\begin{equation}
    \hat{X}_t\propto p(Z_t\mid X_t,M)p(X_t|X_{t-1},U_t)p(X_{t-1}|Z_{1:t-1},U_{1:t-1}),
\end{equation}
where $p(Z_t\mid X_t,M)$ is the observation model, $p(X_t|X_{t-1},U_{1:t})$ is the motion model, and $p(X_{t-1}|Z_{1:t-1},U_{1:t-1})$ is the prior information of the historical poses.  This formulation is also known as recursive filtering for localization, and one representative method for this task is MCL \citep{Global_LiDAR_Loc_Survey}. Employing MCL for initial localization and subsequently leveraging point cloud registration to achieve local pose estimation is a common approach to achieve accurate localization \citep{LocNet}.

\subsection{Monte Carlo Localization}\label{subsec_mcl}
MCL, also known as particle filter localization, is a classical method in robot localization and navigation, achieving localization without initial poses \citep{canedo2016particle, guan2019kld}. The key idea of MCL is to maintain the probability density of the system's pose $X_{t}$ at time $t$ by a group of particles $P=\{P_i|P_i=(x_i,y_i,yaw_i)\}$:
\begin{equation}
\begin{aligned}p(X_t|Z_{1:t},U_{1:t})=\eta p(Z_t|X_t,M)\cdot\\\int p(X_t|X_{t-1},U_t)p(X_{t-1}|Z_{1:t-1},U_{1:t-1})~dX_{t-1},\end{aligned}
\end{equation}
where $\eta$ is a normalization constant obtained from Bayes' rule. MCL includes the steps of particle initialization, state prediction, weight updating, and particle resampling. The observation model $p(Z_{t} |X_{t},M)$ directly determines the reasonability and reliability of the particle weight updating, and it is the most critical factor affecting the localization performance.

\subsection{Spectral matching}\label{subsec_preliminary_sm}
Spectral matching \citep{Spectral_matching} is an efficient spectral method for finding consistent correspondences between two sets of features and can be used for point cloud registration. Given the initial feature correspondences $C_{Init}=\{(a_i,b_i)\}$, an affinity matrix $A\in\mathbb{R}^{n\times n}$ can be constructed, where $A_{i,j}$ is the affinity of $i$th correspondence $(a_i,b_i)$ and $j$th correspondence $(a_j,b_j)$, and the affinity measures how well two correspondences match with each other. The goal of finding inlier correspondences from the initial correspondences can be formulated as finding a cluster $C_{SM}\subseteq C_{Init}$ of correspondences such that the corresponding inter-cluster score $S$ is maximized, i.e:
\begin{equation}\label{eq_inter_cluster_score}
    S=\sum_{(a_i,b_i),(a_j,b_j)\in C_{SM}}A_{i,j}=V^TAV,
\end{equation}
where $V$ is the state vector to represent $C_{SM}$, and $V(a_i,b_i )$ is equal to 1 when $(a_i,b_i)\in C_{Init}$, and 0 otherwise. The optimal solution $V^*$ of Eq. (\ref{eq_inter_cluster_score}) can be calculated by:
\begin{equation}\label{eq_sgv_opt_solution}
    V^*=\operatorname{argmax}(V^TAV).
\end{equation}
As demonstrated by \citet{Spectral_matching},  the approximated optimal solution $V^*$ can be calculated as the principal eigenvector of $A$ by the Raleigh’s ratio theorem \citep{parlett1998symmetric}. Then the inter-cluster score corresponding to the optimal solution $S^*$ can be calculated by:
\begin{equation}\label{eq_sgv_opt_score}
    S^*=V^*{}^TAV^*.
\end{equation}

\begin{figure*}[htbp]
\centering
\includegraphics[width=1.0\textwidth]{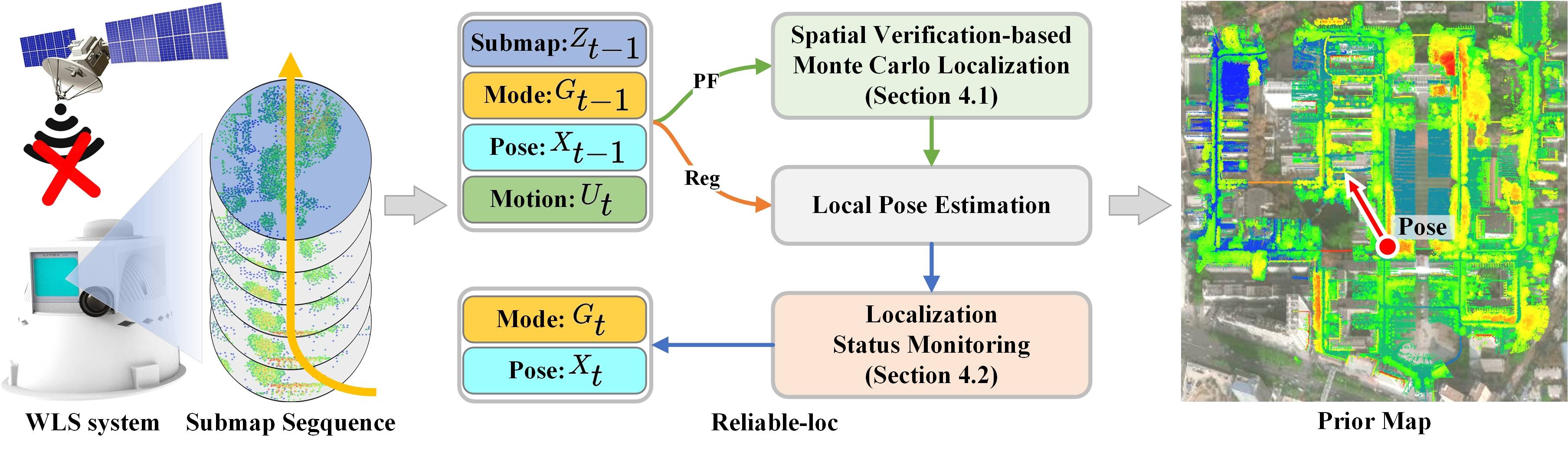}
\caption{Overview of the proposed approach.}\label{fig_reliable_loc}
\end{figure*}

\subsection{Verifiable cues}
Few studies focus on localization quality evaluation and recovery from localization failures, which is important for localization robustness \citep{Self_adaptive_MCL, Yin_A_failure_detection}. In this paper, we look for cues that can verify the localization's quality or reliability so that we can avoid even recovery from localization failures. This leads to the notion of \emph{verifiable cues}, i.e. cues that can verify localization results.

Specifically, for a query point cloud and a point cloud in the prior map, certain spatial properties such as distances and angles are preserved between them. Such spatial invariances, i.e. \emph{spatially verifiable cues}, can be used to assess the similarity of scenes \citep{SGV} and verify the reliability of the local pose estimation \citep{Almqvist_Learning}. For a point cloud sequence, the sequential motions and uncertainties \citep{wan2021enhance}, i.e. \emph{temporal verifiable cues}, provided by the odometry can also verify the reliability of the current localization result. By leveraging both spatial and temporal verifiable cues, we can enhance localization robustness by dynamically adapting the localization approach in various and complex scenes.

\section{Methodology}\label{sec_method}
To achieve robust and accurate global localization, we propose an MCL-based method using spatially verifiable cues (Section \ref{subsec_mcl_gv}) and a localization status monitoring mechanism using temporal verifiable cues (Section \ref{subsec_status_monitor}), as shown in Figure \ref{fig_reliable_loc}. The spatial verification-based MCL designs an observation model utilizing the rich information embedded in local features. The pose uncertainty-based localization status monitor mechanism assesses the reliability of localization results and adaptively switches the localization mode.

\subsection{ Spatial verification-based Monte Carlo Localization}\label{subsec_mcl_gv}
MCL is usually integrated with place recognition techniques to achieve global localization in large-scale scenes, such as Overlap-loc \citep{overlap_loc} and LocNet \citep{LocNet}. Most methods only use global features to update the particle weights. In feature-insufficient scenes, the discrimination of global features decreases dramatically, posing a great challenge to localization robustness. While many point cloud place recognition methods, such as PatchAugNet \citep{PatchAugNet}, LCDNet \citep{Lcdnet}, and EgoNN \citep{Egonn}, can extract both global and local features, the local features contain richer information and can theoretically be used to adjust particles' weights.

\begin{figure}
    \centering
    \includegraphics[width=1\linewidth]{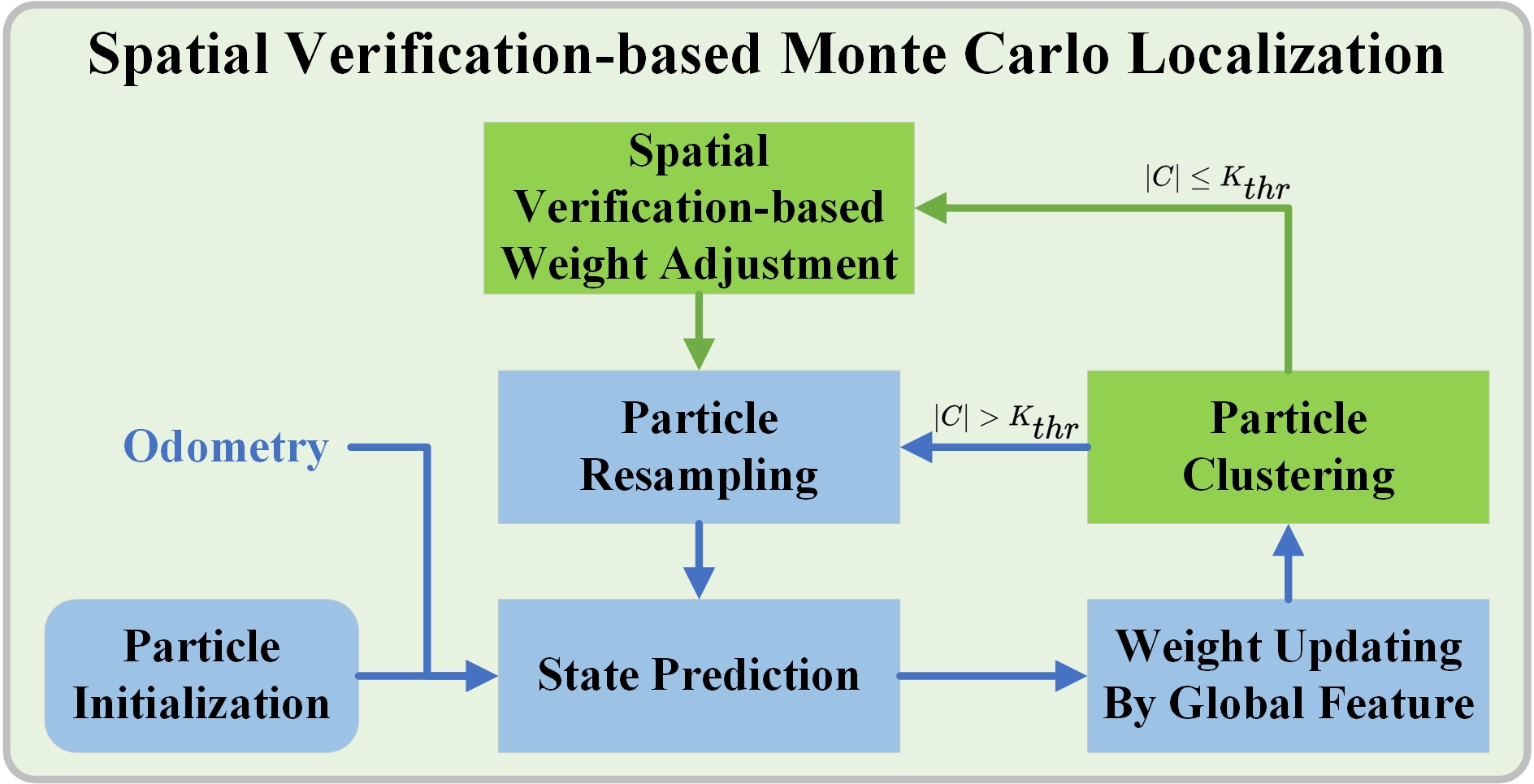}
    \caption{Overview of the spatial verification-based MCL.}
    \label{fig_gv_mcl}
\end{figure}

In this paper,  we propose an MCL-based method using spatially verifiable cues through spatial verification, as shown in Fig.\ref{fig_gv_mcl}. The proposed method uses both global and local features from place recognition to construct the observation model, which consists of two parts: global feature-based $p_{G}(Z_t|X_t,M)$, and spatial verification-based. The global feature-based observation model is presented in Remark \ref{remark1}. The spatial verification-based observation model consists of two parts: spectral matching-based $p_{SM}(Z_t|X_t,M)$ and pose error-based $p_{PE}(Z_t|X_t,M)$. Namely, the observation model is formulated as:
\begin{equation}
    p(Z_t|X_t,M)=p_{G}(Z_t|X_t,M)p_{SM}(Z_t|X_t,M)p_{PE}(Z_t|X_t,M).
\end{equation}

\textbf{Particle clustering}. Due to the large quantity, the particles are first clustered to improve the efficiency of subsequent spatial verification-based observation model construction. Particle clustering is based on the Euclidean distance between particles and can be formulated as:
\begin{equation}
\begin{aligned}C=\{C_k|d\big(P_i,P_j\big)&<d_{cluster},\forall P_i,P_j\in C_k;\\d\big(P_i,P_j\big)&>d_{cluster},\forall P_i\in C_k,\forall P_j\in C_{n\neq k}\},\end{aligned}
\end{equation}
where $P_i$ is the $i$th particle in $P$, $d{\left(P_i,P_j\right)}$ is the Euclidean distance between two particles and $d_{cluster}$ is the threshold to determine whether two particles belong to the same cluster. In general, $d_{cluster}$ is set as the resolution of the map grid.

\begin{remark}\label{remark1}
Global feature-based observation model: given a particle $P_i$ and the closest submap $M_i$ to it in the map, the global feature-based observation model is:
\begin{equation}
    p_G(Z_t|X_t,M)\propto F_G(Z_t,M_i)=\frac{\cos(f_{Z_t},f_{M_i})+1}2,
\end{equation}
Where $F_G$ extracts the global features of $Z_t$ and $M_i$ and calculates their similarity, describing the overall similarity of the scenes where the system and the particles are located. $f_{Z_t}$ and $f_{M_i}$ are the global features of $Z_t$ and $M_i$, respectively. In general, the information embedded in the global feature is very limited and insufficient to ensure that particles converge in the correct direction, especially in feature-insufficient scenes.
\end{remark}

\textbf{Spatial verification-based weight adjustment}. By introducing the inter-cluster score in spectral matching \citep{Spectral_matching}, we can quantitatively assess the spatial consistency of two point cloud submaps, and thus adjust the particles' weights. 

\begin{figure}
    \centering
    \includegraphics[width=1\linewidth]{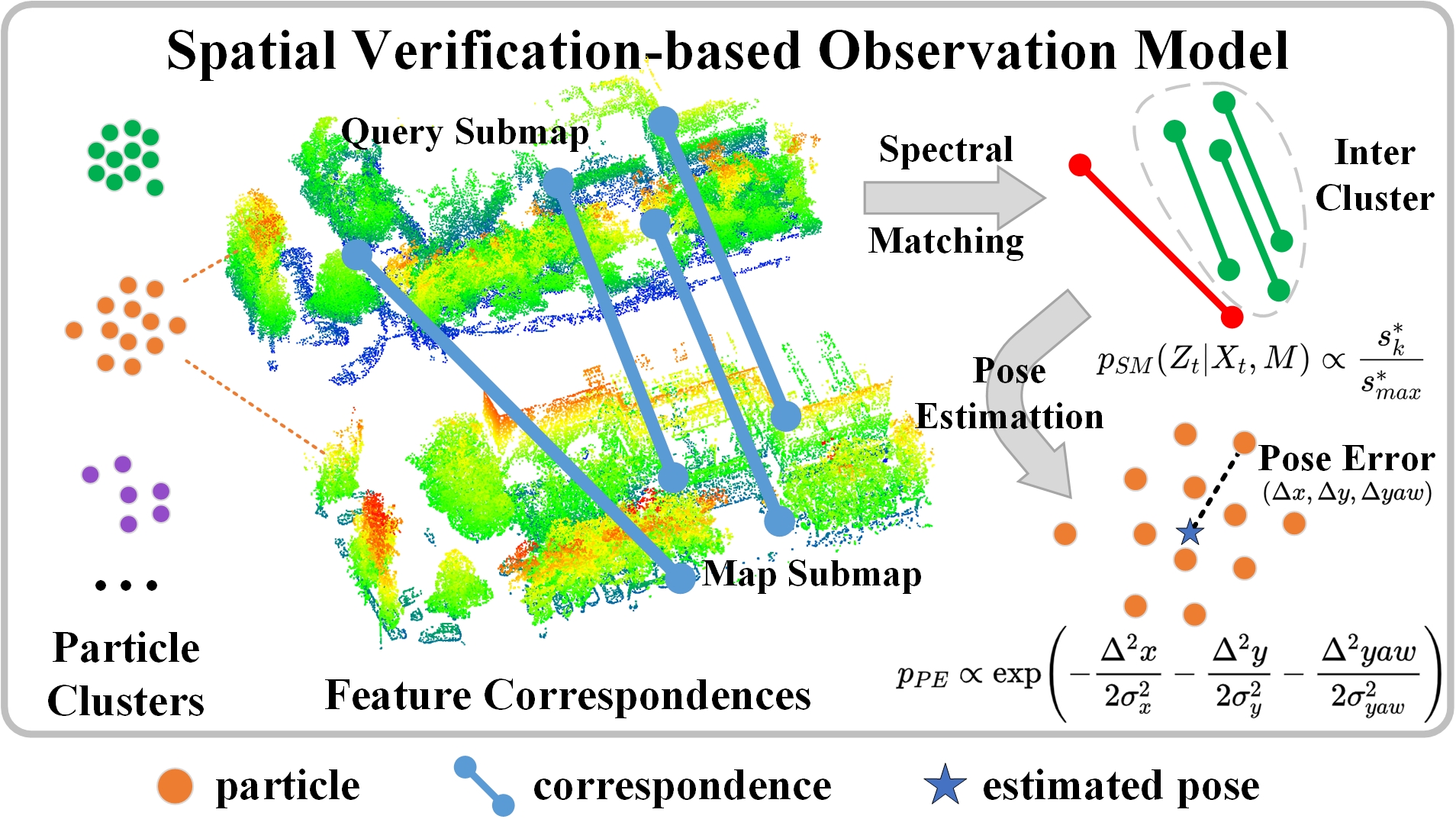}
    \caption{Schema of spatial verification-based observation model.}
    \label{fig_gv_obs_model}
\end{figure}

As shown in Fig.\ref{fig_gv_obs_model}, for each particle cluster $C_k$ and the closest submap $M_k$ to it in the map, the initial correspondences can be obtained by nearest neighbor matching using the local features from place recognition, and the affinity matrix $A\in\mathbb{R}^{n\times n}$ can also be constructed by referring to SGV \citep{SGV}. Then the inter-cluster score $S^*$ corresponding to $Z_t$ and $M_k$ can be calculated according to Subsection \ref{subsec_preliminary_sm}, and the spectral matching-based observation model is:
\begin{equation}
    p_{SM}(Z_{t}|X_{t},M)\propto\frac{s_{k}^{*}}{s_{max}^{*}}, ~ S_{max}^*=\max_mS_m^*,
\end{equation}
where $S_{max}^{*}$ is the maximum of the inter-cluster scores for all particle clusters, $m$ is the index of the particle cluter. It should be noted that all particles in $C_k$ use the same observation model to adjust their weights. Meanwhile, the 3-DOF pose of the system $X_{t}^{*}=\{x_{t}^{*},y_{t}^{*},yaw_{t}^{*}\}$ can be estimated based on the correspondences corresponding to $S^*$ using SVD (Singular Value Decomposition) \citep{SVD}.

For each particle $P_i$ in cluster $C_k$, the pose error-based observation model can be constructed by the difference between the estimated pose $X_{t}^{*}$ and the state $\{x_i,y_i,yaw_i\}$ of $P_i$ as:
\begin{equation}\label{eq_pose_error_obs_model}
\begin{gathered}p_{PE}(Z_t|X_t,M)\propto\\\exp\left(-\frac{(x_i-x_t^*)^2}{2\sigma_x^2}-\frac{(y_i-y_t^*)^2}{2\sigma_y^2}-\frac{(yaw_i-yaw_t^*)^2}{2\sigma_{yaw}^2}\right),\end{gathered}
\end{equation}
where $\sigma_x$, $\sigma_y$, and $\sigma_{yaw}$ control the range of the Gaussian kernel function in each dimension, respectively. It should be noted that the spatial verification-based weight adjustment is only used when the number of particle clusters is less than $K_{thr}$.

\subsection{Localization status monitoring}\label{subsec_status_monitor}
\begin{figure}
    \centering
    \includegraphics[width=0.7\linewidth]{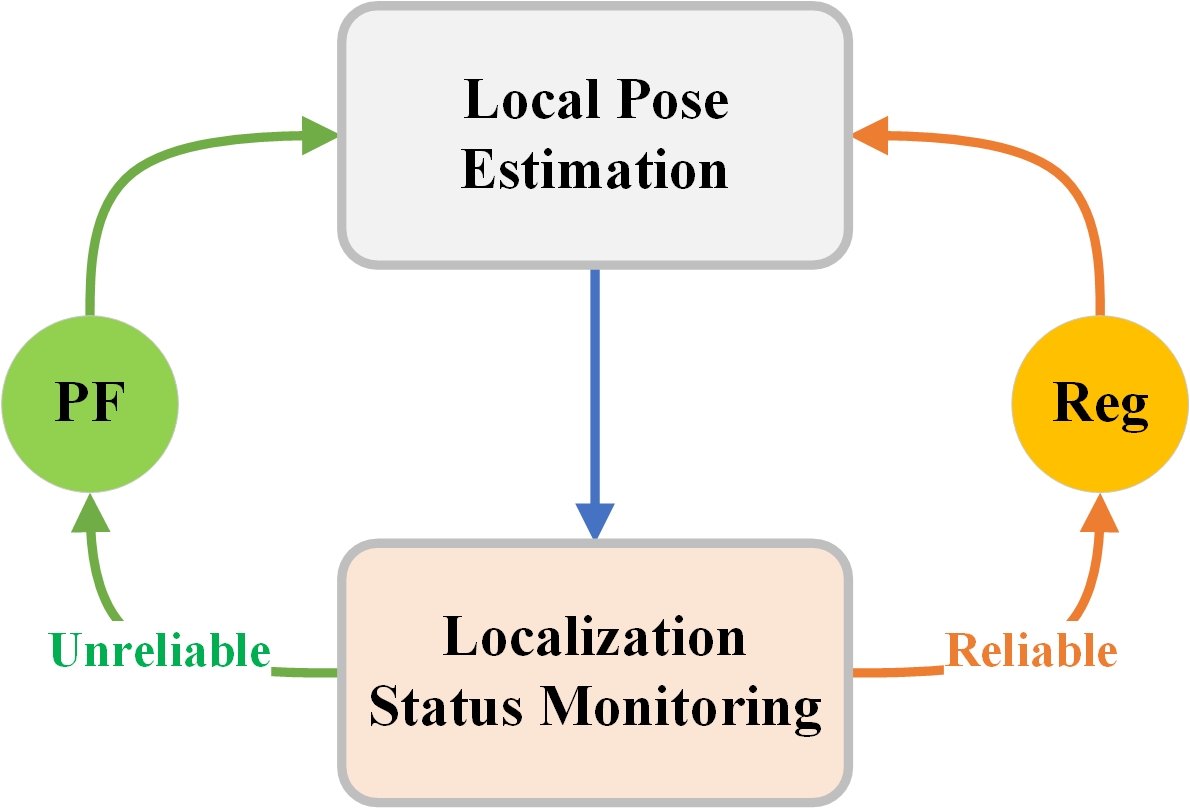}
    \caption{Schema of localization mode switching.}
    \label{fig_loc_mode_switching}
\end{figure}

Several methods rely on local pose estimation to achieve accurate localization once MCL converges \citep{LocNet, Seqpolar}, but they are not robust in scenes with insufficient features and incomplete map coverage. To address this problem, we propose a localization status monitoring mechanism using verifiable cues from spatial and temporal aspects, i.e. pose uncertainty from feature correspondences and odometry, which assesses the reliability of localization results and adaptively switches the localization mode. As shown in Fig.\ref{fig_loc_mode_switching}, if the current localization result is reliable, the local pose estimation will be used for subsequent localization to ensure localization accuracy; If the current localization result is unreliable, the spatial verification-based MCL will be used for subsequent localization to ensure localization robustness. For convenience, we denote the localization mode as $G$, the localization based on point cloud registration as $Reg$, and the localization based on MCL as $PF$.

\begin{figure}
    \centering
    \includegraphics[width=0.775\linewidth]{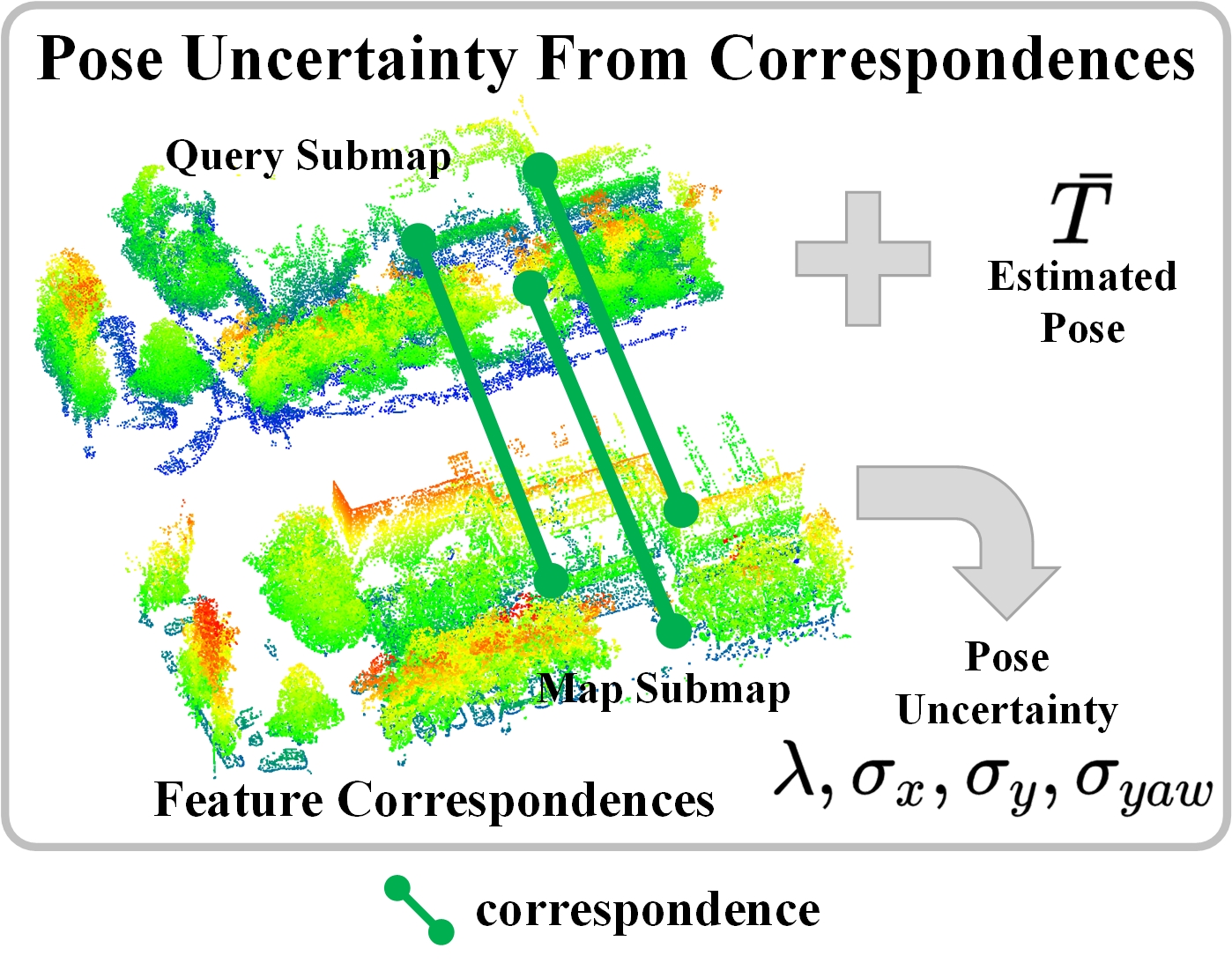}
    \caption{Schema of pose uncertainty from correspondences.}
    \label{fig_pose_uncertainty_corr}
\end{figure}

\begin{figure}
    \centering
    \includegraphics[width=0.70\linewidth]{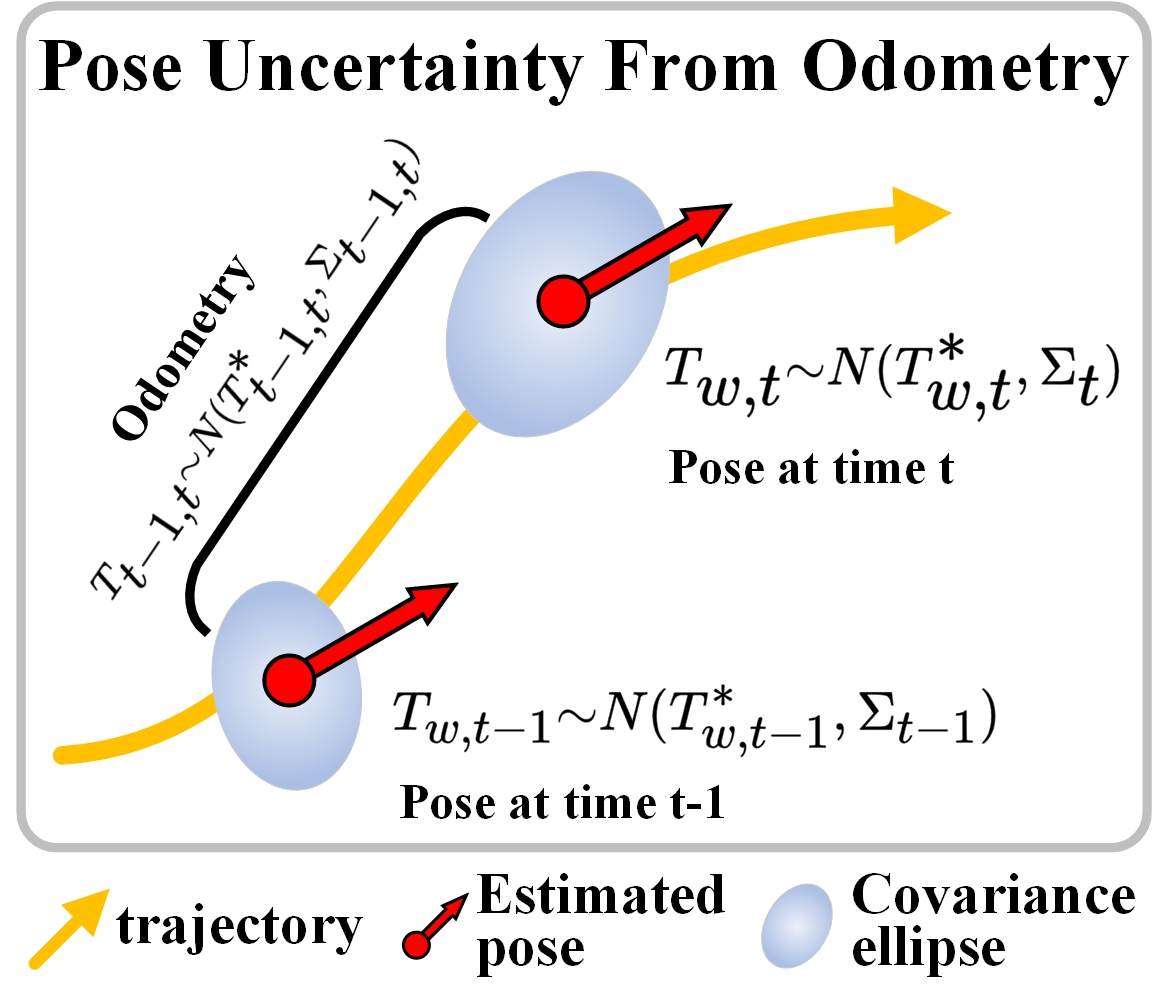}
    \caption{Schema of pose uncertainty from odometry.}
    \label{fig_pose_uncertainty_odo}
\end{figure}

Pose covariance reflects the uncertainty of the estimated pose in different dimensions, and it's a common indicator of pose uncertainty. Besides, the minimum eigenvalue of the Hessian matrix based on feature correspondences for local pose estimation can also reflect the pose uncertainty holistically.  Both of them can be used as a basis for monitoring the localization status. In Reliable-loc, the pose uncertainty can be obtained from feature correspondences (spatially verifiable cues) in local pose estimation and odometry (temporal verifiable cues), as shown in Fig.\ref{fig_pose_uncertainty_corr} and Fig.\ref{fig_pose_uncertainty_odo}.

\textbf{Pose uncertainty from correspondences}. For local pose estimation, we first find reliable feature correspondences through spectral matching \citep{Spectral_matching}, and then use Teaser++ \citep{Teaser} to solve the 4-DOF pose $T_{w,t-1}$ of the system, and finally compute the standard deviation of the pose in $x$, $y$, and $yaw$ dimensions:
\begin{equation}
    P_w=T_{w,t-1}\cdot P_{t-1},T_{w,t-1}{\sim}N(\bar{T}_{w,t-1},\mathit{\Sigma}_{t-1}),
\end{equation}
\begin{equation}
    \left(\sigma_{x}, \sigma_{y}, \sigma_{\text {yaw }}\right)=\operatorname{sqrt}\left(\operatorname{diag}\left(\mathit{\Sigma}_{t-1}\right)\right)_{[0,1,3]},
\end{equation}
Where $J_{t-1}$ is the Jacobi matrix computed from the feature correspondences at time $t-1$, $\mathit{\Sigma}_{t-1}=(J_{t-1}^{T}\cdot J_{t-1})^{-1}$ is the pose covariance, $P_{t-1}$ and $P_w$ are the coordinates of the feature points in the query and map submaps, $[.]_{[i]}$ is $i$th element of the vector. We further compute the minimum eigenvalue of the Hessian matrix:
\begin{equation}
    \lambda=\min\left(\operatorname{eig}(J_{t-1}^T\cdot J_{t-1})\right).
\end{equation}
It indicates the reliability of the pose estimation holistically, and the larger its value the more reliable the pose estimation is.

\textbf{Pose uncertainty from odometry}. In scenes with insufficient features or incomplete map coverage, local pose estimation is unreliable or can't work, and we have to use odometry to derive the system's pose and the corresponding uncertainty. Given the pose $T_{w,t-1}{\sim}N(T_{w,t-1}^{*},\mathit{\Sigma}_{t-1})$ at time $t-1$ and the motion $U_{t}\doteq T_{t-1,t}{\sim}N(T_{t-1,t}^{*},\mathit{\Sigma}_{t-1,t})$ provided by odometry from time $t-1$ to time $t$, the pose at time $t$ is:
\begin{equation}
    T_{w,t}=T_{w,t-1}\cdot T_{t-1,t}\sim N(T_{w,t}^{*},\mathit{\Sigma}_{t}),
\end{equation}
where $\left.\left\{\begin{matrix}T_{w,t}^*=T_{w,t-1}^*\cdot T_{t-1,t}^*\\\mathit{\Sigma}_t\doteq\mathit{\Sigma}_{t-1}+J_{t-1,t}^T\cdot\mathit{\Sigma}_{t-1,t}\cdot J_{t-1,t}\end{matrix}\right.\right.$, and $(\sigma_x,\sigma_y,\sigma_{yaw})=sqrt(diag(\mathit{\Sigma}_t))_{[0,1,3]}$ is the corresponding standard deviation.

\textbf{Localization status}. Based on the above pose uncertainty, the localization status can be obtained as follows:
\begin{equation}  
    H = \left\{  
    \begin{array}{l}  
        \text{unreliable,} \quad \text{if } \lambda < \lambda_{thr} \text{ or } \sigma_x > \sigma_{x_{thr}} \text{ or } \\  
        \qquad\qquad\qquad \sigma_y > \sigma_{y_{thr}} \text{ or } \sigma_{yaw} > \sigma_{yaw_{thr}}, \\  
        \text{reliable,} \qquad\text{ else}  
    \end{array}  
    \right.  
\end{equation}
Where $\lambda_{thr}$, $\sigma_{x_{thr}}$, $\sigma_{y_{thr}}$, $\sigma_{yaw_{thr}}$ are the thresholds of these indicators. 'reliable' means the estimated pose is reliable, and 'unreliable' is the opposite.

\textbf{Localization mode switching}. In the initial stage, coarse localization in the large-scale prior map is achieved through MCL, namely, the initial localization mode is $PF$. After MCL converges, the optimal localization mode is automatically selected based on the localization statuses, thus ensuring localization performance. As shown in Fig.\ref{fig_loc_mode_switching}, localization mode switching consists of the following three cases:
\begin{enumerate}[1)]
    \item $H==reliable$: the subsequent localization will be performed in $Reg$ mode.
    \item $G==PF$ and $H==unreliable$: the subsequent localization will be performed in $PF$ mode.
    \item  $G==Reg$ and $H==unreliable$: the localization mode will be switched to $PF$, and the particles will be reinitialized according to the estimated pose and covariance.
\end{enumerate}

\textbf{Particle re-initialization}. In case 3), the particles are reinitialized as follows:
\begin{equation}\label{eq_particle_reinit}
    \left.\left\{\begin{matrix}x_i\boldsymbol{\sim}N(x_i^*,(\alpha\sigma_x)^2)\\y_i\boldsymbol{\sim}N(y_i^*,(\alpha\sigma_y)^2)\\yaw_i\boldsymbol{\sim}N(yaw_i^*,(\alpha\sigma_{yaw})^2)\\w_i=\exp{(-\frac{(x_i-x_t^*)^2}{2\sigma_x^2}-\frac{(y_i-y_t^*)^2}{2\sigma_y^2}-\frac{(yaw_i-yaw_t^*)^2}{2\sigma_{yaw}^2})}\end{matrix}\right.\right.,
\end{equation}
where $(x_i^*,y_i^*,yaw_i^*)$ is the system's pose, $(\sigma_x,\sigma_y,\sigma_{yaw})$ is the corresponding standard deviation, $(x_{i},y_{i},yaw_{i},w_{i})$ is the state and weight of the particle, and $\alpha$ is the coefficient.

\section{Experiment}\label{sec_experiment}
\subsection{Experiment data}\label{subsec_exp_data}
\begin{figure*}[ht]
\centering
\includegraphics[width=1\textwidth]{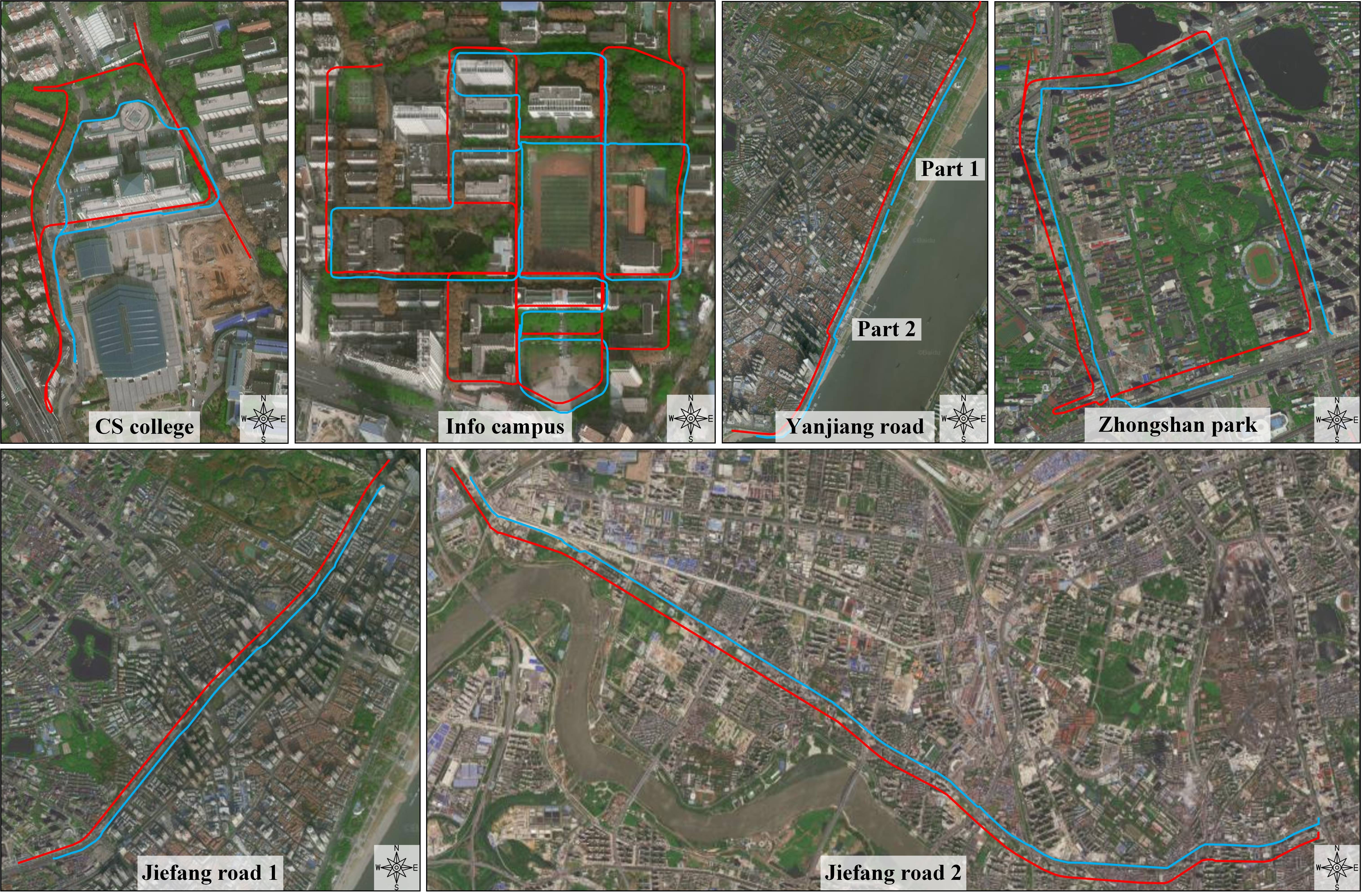}
\caption{Overview of the experimental data: the red and blue lines are the acquisition trajectories of the vehicle-mounted MLS system and the helmet-mounted WLS system respectively (manually offset for display purposes).}\label{fig_exp_data}
\end{figure*}

In this paper, the proposed method's effectiveness is verified on a heterogeneous point cloud dataset collected by a vehicle-mounted MLS system and a helmet-mounted WLS system in Wuhan, China. High-precision vehicle-mounted MLS point clouds serve as the prior map and helmet-mounted WLS point clouds are used for localization. The dataset contains seven data: CS college, Info campus, Zhongshan park, Jiefang road 1, Jiefang road 2, Yanjiang road 1, Yanjiang road 2, as shown in Fig.\ref{fig_exp_data}.
The first two data are collected on a college campus with abundant vegetation, where roadside buildings are often obscured by trees. The road of CS college is relatively open, whereas the Info campus exhibits a more unstructured environment, with numerous blank data holes present in the prior map. The last five data are collected on urban roads with regularly arranged roadside trees, lots of high-rise buildings and viaducts in the scene, and lots of dynamic objects on the road, with variations in some of the scenes. Some of the road sections in the last five data have insufficient features, posing a significant challenge for localization. The details of creating the dataset are described in our previous work, PatchAugNet \citep{PatchAugNet}. It should be noted that WLS point clouds of Jiefang Road 2 is captured with Livox Mid 360, which is significantly different in scanning modes. The difference lies in the fact that the helmet point cloud submaps are generated every 0.5 m along the trajectory and the MLS point cloud submaps are derived by utilizing $5\times5$ m regular grids. Additionally, we introduce noise into the helmet trajectory by referencing the official code of Overlap-loc \citep{overlap_loc}. The comprehensive details of the experimental data are presented in Table \ref{table_exp_data}.

\begin{table*}[htbp]
\centering
\fontsize{7}{10}\selectfont
\caption{Detailed description of experimental data.}
\label{table_exp_data}
\begin{tblr}{
  width = \linewidth,
  colspec = {Q[125]Q[200]Q[81]Q[81]Q[81]Q[71]Q[71]Q[79]Q[67]},
  cells = {c},
  cell{1}{1} = {r=2}{},
  cell{1}{2} = {r=2}{},
  cell{1}{3} = {r=2}{},
  cell{1}{4} = {c=2}{0.194\linewidth},
  cell{1}{6} = {c=2}{0.142\linewidth},
  cell{1}{8} = {r=2}{},
  cell{1}{9} = {r=2}{},
  cell{3}{4} = {r=7}{},
  cell{3}{5} = {r=2}{},
  cell{3}{7} = {r=2}{},
  cell{5}{2} = {r=5}{},
  cell{5}{5} = {r=5}{},
  cell{8}{9} = {r=2}{},
  hline{1,3,10} = {-}{},
}
\textbf{Datasets}        & \textbf{Scene description}                                       & \textbf{Difficult level} & \textbf{Acquisition equipment} &                      & \textbf{Acquisition time} &              & {\textbf{Query}\\\textbf{trajectory}\\\textbf{length(km)}} & {\textbf{Number}\\\textbf{of map}\\\textbf{submaps}} \\
                         &                                                                  &                          & \textbf{Query}                 & \textbf{Map}         & \textbf{Query}            & \textbf{Map} &                                                            &                                                      \\
\textbf{CS college}      & University campus road.                                          & Easy                     & WHU-Helmet                     & CHCNAV Alpha 3D      & 2022-07-29                & 2021-11-20   & 0.96                                                       & 2493                                                 \\
\textbf{Info campus}     & University campus road, partial map is incomplete.               & Hard                     &                                &                      & 2021-11-12                &              & 3.10                                                        & 5878                                                 \\
\textbf{Zhongshan park}  & Urban road, the features of some road sections are insufficient. & Medium                   &                                & Hi-Target HiScan-VUX & 2022-08-23                & 2019-12-11   & 4.74                                                       & 15186                                                \\
\textbf{Jiefang road 1}  &                                                                  & Medium                   &                                &                      & 2022-08-23                & 2020-03-20   & 4.58                                                       & 12202                                                \\
\textbf{Jiefang road 2}  &                                                                  & Hard                     &                                &                      & 2024-12-13                & 2020-03-20   & 11.73                                                      & 30090                                                \\
\textbf{Yanjiang road 1} &                                                                  & Hard                     &                                &                      & 2022-08-23                & 2020-03-22   & 2.86                                                       & 13772                                                \\
\textbf{Yanjiang road 2} &                                                                  & Medium                   &                                &                      & 2022-08-23                & 2020-03-22   & 3.24                                                       &                                                      
\end{tblr}
\end{table*}

\subsection{Experiment setting}\label{subsec_exp_setting}
\textbf{Comparison methods.} We use Overlap-loc \citep{overlap_loc} as the baseline to verify the effectiveness of the proposed method. The comparison methods in the quantitative evaluation experiments include PF-loc, PF-SGV-loc, PF-SGV2-loc, Reg-loc, and Reliable-loc. PF-loc is the baseline method, and its difference from the original Overlap-loc \citep{overlap_loc} lies in the place recognition method used. PF-SGV-loc enhances PF-loc by utilizing the spectral matching-based observation model. PF-SGV2-loc further improves upon PF-SGV-loc by incorporating the pose error-based observation model. Reg-loc employs only the $Reg$ localization mode after MCL converges. Reliable-loc is the proposed method, automatically switching the localization mode based on pose uncertainty.

\textbf{Implementary details.} We use the official model supplied by PatchAugNet \citep{PatchAugNet} to extract global and local features. The spectral matching method described in Subsection \ref{subsec_preliminary_sm} utilizes the official code of SGV \citep{SGV}, renowned for its excellent parallelism capabilities. Furthermore, the local pose estimation in Subsection \ref{subsec_status_monitor} is achieved based on the official code of Teaser++ \citep{Teaser}, renowned for its robust performance. The number of particles initialized for MCL is 5000, while after convergence, it reduces to 400. In particle clustering, the distance threshold $d_{cluster}$ is set to 5 m, and $K_{thr}$ is set to 40. The parameters of the Gaussian kernel function in Eq. (\ref{eq_pose_error_obs_model}) are set to 30 m, 30 m, and 60 degrees, respectively. Low-variance resampling \citep{thrun2002probabilistic} method is used for particle resampling. For localization status monitoring, the thresholds $\sigma_{x_{thr}}$, $\sigma_{y_{thr}}$, and $\sigma_{yaw_{thr}}$ are set to 30 m, 30 m, and 15 degrees, respectively. $\lambda_{thr}$ is assigned a value of 1e-4 on Yanjiang road 1, and 5e-4 for the remaining datasets. All experimental methods are implemented in Python, and all experiments are conducted on an Intel® Core i7-13700KF CPU and an Nvidia® GeForce RTX 3080Ti GPU.

\textbf{Evaluation metrics.} To evaluate the proposed method's effectiveness, we compare the estimated trajectories with the ground truth trajectories and then calculate the mean error and root mean square error (RMSE) of position and yaw angle. We denote $PE$ and $YE$ as position error and yaw error respectively. It should be noted that only results after MCL converges are considered. Additionally, we count the proportion of position errors less than 2, 5, 10, 15, 20, 25, and 30 meters and yaw angle errors less than 2, 5, 10, 15, 20, 25, and 30 degrees, i.e., the localization success rate under different position and yaw error thresholds. For convenience, we denote $R@XmYd$ as the proportion of position errors less than $X$m and yaw errors less than $Y$ degrees. Similarly, $R@Xm$ denotes the proportion of position errors less than $X$m. 

\subsection{Quantitative evaluation}\label{subsec_quanitative_eval}
\begin{figure*}[htbp]
\centering
\includegraphics[width=0.85\textwidth]{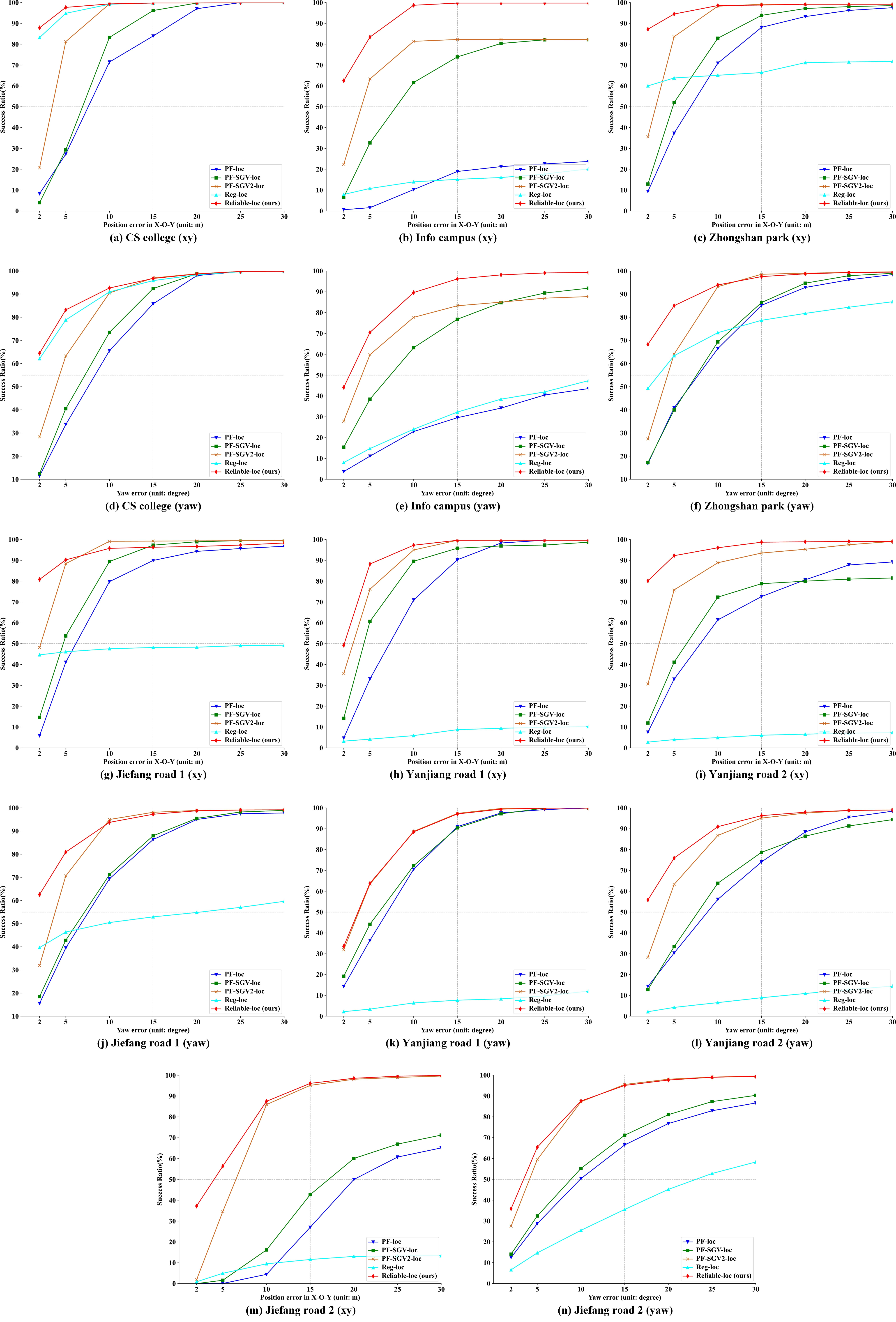}
\caption{Localization success rate curves of each method on experimental data. (a), (b), (c), (g), (h), (i), (m) are the $R@Xm$ curves, and (d), (e), (f), (j), (k), (l), (n) are the $R@Yd$ curves.}\label{fig_quantitative_eval_curve}
\end{figure*}

This section compares and analyzes the global localization performance of PF-loc, PF-SGV-loc, PF-SGV2-loc, Reg-loc, and Reliable-loc on the experimental data. Table \ref{table_quantitative_eval} shows the localization success rate (position error $<2$ m  and yaw error $<5\deg$ ) and localization accuracy of all methods, Fig.\ref{fig_quantitative_eval_curve} shows the localization success rate curves of all methods.

Compared with PF-loc, $R@2m5deg$, position accuracy, and yaw accuracy of PF-SGV-loc are 2.14 percent, 123.00 m, and 5.50 degrees higher, respectively. Similarly, $R@2m5deg$, position accuracy, and yaw accuracy of PF-SGV2-loc are 11.70 percent, ±108.12 m, and ±4.84 degrees higher than that of PF-SGV-loc, respectively. The reason for performance improvement is that the spatial verification-based observation model can effectively utilize the rich information embedded in local features, which is an effective complement to the observation model based on global features. Specifically, in feature-rich scenes, higher localization accuracy can be achieved with the spatial-verification-based observation model, providing a better initial pose for subsequent localization. In feature-insufficient scenes, higher localization robustness can be achieved by avoiding the particles converging to the incorrect region.

PF-SGV2-loc presents a high $R@30m$ on all data, exhibiting excellent robustness. However, the corresponding position and yaw accuracy are likely as low as ±35.15 m and ±11.65 degrees, respectively, exhibiting poor accuracy. Reg-loc achieves position and yaw accuracies of ±1.01 m and ±2.51 degrees respectively on the simple data (CS college), exhibiting excellent accuracy. While it presents a low $R@30m$ in other data with insufficient features, exhibiting poor robustness. Reliable-loc outperforms both PF-SGV2-loc and Reg-loc on all data with a total position accuracy and yaw accuracy of ±2.91 m and ±3.74 degrees, respectively, and it also presents a high $R@30m$ on all data, exhibiting excellent robustness and accuracy. The experimental results show that MCL is robust but not accurate, and localization based on point cloud registration is effective in feature-rich scenes but prone to failure in scenes with insufficient features and incomplete map coverage. However, Combining the two methods significantly improves the localization robustness and accuracy. The reasons are as follows: 1) the particles in MCL possess certain exploratory capabilities, allowing them to converge to the correct region even when the initial pose is inaccurate.; 2) After MCL converges, local pose estimation can achieve high localization accuracy in scenes with sufficient distinguishable features and provide an accurate initial pose for subsequent localization; 3) Reliable-loc can improve localization performance by adaptively switching the localization mode based on pose uncertainty from correspondences and odometry. It uses the $Reg$ localization mode when the initial pose is accurate and there are abundant features in the scene while using the $PF$ localization mode in other situations.

\begin{table*}[htbp]
\centering
\fontsize{5.8}{10}\selectfont
\caption{Localization success rate and accuracy of all methods on experimental data.}
\label{table_quantitative_eval}
\begin{tblr}{
  width = \linewidth,
  colspec = {Q[83]Q[100]Q[77]Q[92]Q[96]Q[92]Q[96]Q[94]Q[102]Q[96]},
  cell{2}{1} = {r=3}{},
  cell{5}{1} = {r=3}{},
  cell{8}{1} = {r=3}{},
  cell{11}{1} = {r=3}{},
  cell{14}{1} = {r=3}{},
  hline{1-2,5,8,11,14,17} = {-}{},
}
\textbf{Methods}                         & \textbf{Metrics}      & \textbf{CS college}  & \textbf{Info campus} & \textbf{Zhongshan park} & \textbf{Jiefang road 1} & \textbf{\textbf{Jiefang road 2}} & \textbf{Yanjiang road 1} & \textbf{Yanjiang road 2} & \textbf{Total}       \\
\textbf{PF-loc}                          & \textbf{R@2m5d (\%)↑} & 4.76                 & 0.39                 & 5.14                    & 1.50                    & 0.00                             & 1.50                     & 4.29                     & 1.87                 \\
                                         & \textbf{PE (m)↓}      & 8.96 ± 10.41         & 241.05 ± 289.59      & 8.28 ± 14.43            & 7.77 ± 10.49            & 556.14 ± 1628.78                 & 7.65 ± 8.91              & 12.12 ± 17.06            & 238.90 ± 1011.32     \\
                                         & \textbf{YE (deg)↓}    & 7.98 ± 9.49          & 53.26 ± 67.94        & 8.25 ± 10.73            & 7.90 ± 10.48            & 17.12 ± 30.83                    & 7.64 ± 9.24              & 9.90 ± 12.26             & 15.72 ± 28.87        \\
\textbf{PF-SGV-loc}                      & \textbf{R@2m5d (\%)↑} & 0.81                 & 2.33                 & 7.00                    & 9.24                    & 0.00                             & 6.58                     & 7.96                     & 4.01                 \\
                                         & \textbf{PE (m)↓}      & 7.16 ± 7.95          & 23.49 ± 76.67        & 6.34 ± 12.92            & 5.32 ± 6.39             & 288.96 ± 1137.59                 & 5.38 ± 7.27              & 12.66 ± 20.43            & 115.90 ± 701.82      \\
                                         & \textbf{YE (deg)↓}    & 6.68 ± 7.96          & 10.91 ± 22.78        & 7.92 ± 10.31            & 7.31 ± 9.34             & 13.15 ± 21.83                    & 6.96 ± 8.80              & 9.83 ± 13.29             & 10.22 ± 16.94        \\
\textbf{PF-SGV2-loc}                     & \textbf{R@2m5d (\%)↑} & 11.25                & 17.77                & 23.11                   & 34.75                   & 1.16                             & 27.00                    & 24.45                    & 15.71                \\
                                         & \textbf{PE (m)↓}      & 3.38 ± 3.73          & 35.15 ± 102.23       & 3.22 ± 10.71            & 2.57 ± 3.16             & 6.98 ± 8.09                      & 3.31 ± 4.11              & 4.43 ± 6.73              & 7.78 ± 32.37         \\
                                         & \textbf{YE (deg)↓}    & 4.17 ± 5.20          & 11.65 ± 28.36        & 4.38 ± 5.58             & 3.80 ± 4.84             & 5.20 ± 7.15                      & 4.53 ± 5.97              & 4.91 ± 6.64              & 5.38 ± 10.64         \\
\textbf{Reg-loc}                         & \textbf{R@2m5d (\%)↑} & 73.78                & 6.64                 & 55.37                   & 42.95                   & 0.63                             & 2.71                     & 2.00                     & 18.85                \\
                                         & \textbf{PE (m)↓}      & 1.01 ± 1.83          & 164.37 ± 201.16      & 113.08 ± 227.61         & 458.15 ± 773.78         & 594.66 ± 786.47                  & 762.37 ± 938.22          & 1356.16 ± 1579.88        & 529.77 ± 820.08      \\
                                         & \textbf{YE (deg)↓}    & 2.51 ± 4.37          & 32.22 ± 39.78        & 9.82 ± 18.26            & 34.25 ± 54.00           & 33.52 ± 46.20                    & 65.78 ± 75.52            & 58.02 ± 63.06            & 33.42 ± 47.49        \\
{\textbf{Reliable-loc}\\\textbf{(ours)}} & \textbf{R@2m5d (\%)↑} & \textbf{77.55}       & \textbf{52.02}       & \textbf{80.39}          & \textbf{72.37}          & \textbf{30.87}                   & \textbf{35.91}           & \textbf{68.87}           & \textbf{53.16}       \\
                                         & \textbf{PE (m)↓}      & \textbf{0.73 ± 1.10} & \textbf{2.41 ± 3.31} & \textbf{1.13 ± 10.29}   & \textbf{1.92 ± 5.34}    & \textbf{4.93 ± 6.79}             & \textbf{2.39 ± 3.09}     & \textbf{1.33 ± 2.59}     & \textbf{2.91 ± 6.49} \\
                                         & \textbf{YE (deg)↓}    & \textbf{2.09 ± 3.55} & \textbf{4.02 ± 6.12} & \textbf{2.44 ± 4.57}    & \textbf{2.64 ± 4.42}    & \textbf{4.78 ± 7.05}             & \textbf{4.45 ± 6.02}     & \textbf{3.22 ± 5.23}     & \textbf{3.74 ± 5.91} 
\end{tblr}
\emph{Note: the rightmost column 'Total' is the overall localization performance of each method on all experimental data.}
\end{table*}

\section{Analysis and discussion}\label{sec_analysis_discussion}
In this section, detailed analyses of Reliable-loc are presented in terms of the effectiveness of the spatial verification-based observation model, the necessity of switching the localization mode, the distribution of the localization mode, parameter analysis, efficiency analysis, and failure cases. In addition, an outlook for future work is provided in conjunction with the experimental results and analysis.

\subsection{Effectiveness of the spatial verification-based observation model}\label{subsec_effectiveness_obs_model_gv}
\begin{figure*}[t]
\centering
\includegraphics[width=1.0\linewidth]{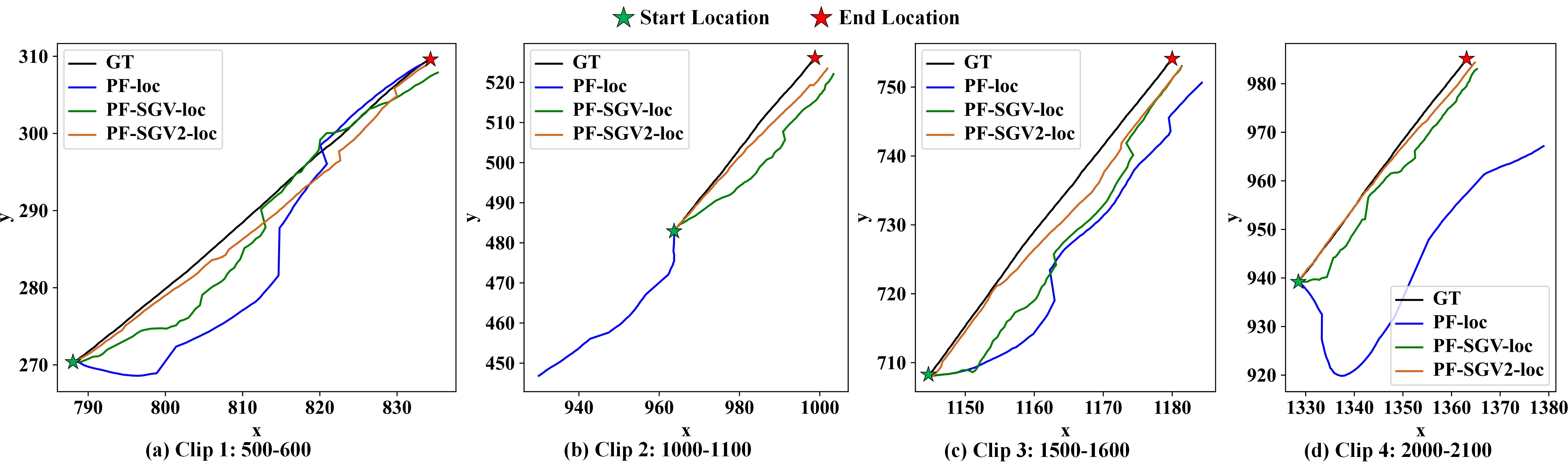}
\caption{Localization trajectories in the four clips when the initial position is known and the yaw is unknown. The black, blue, green, and orange lines are the trajectories of the Ground Truth, PF-loc, PF-SGV-loc, and PF-SGV2-loc, respectively, with the green pentagram as the starting point and the red pentagram as the end point.}\label{fig_effectiveness_obs_model_gv}
\end{figure*}

To further demonstrate the effectiveness of the spatial verification-based observation model, we extracted four 100-frame clips from the experimental data of Jiefang road 1: 500-600, 1000-1100, 1500-1600, and 2000-2100. We then compared the localization performance of PF-loc, PF-SGV-loc, and PF-SGV2-loc on these clips. The particles are initialized with a known position but an unknown yaw in the experiment. The experimental results are presented in Fig.\ref{fig_effectiveness_obs_model_gv}.

As shown in Fig.\ref{fig_effectiveness_obs_model_gv}, the localization performance of PF-loc, PF-SGV-loc, and PF-SGV2-loc is improved sequentially on all four clips, and the improvement is most significant from PF-loc to PF-SGV-loc. In relatively simple clip 1 and clip 3, PF-loc's localization accuracy exceeds 10 m. The spatial verification-based observation model substantially improves the localization accuracy. In Clip 2, the particles of PF-loc converge in the wrong direction, whereas the particles of PF-SGV-loc converge correctly using the spectral matching-based observation model. PF-SGV2-loc further adjusts the particles' weights using the pose error-based observation model, ensuring the particles converge to the correct region. In Clip 4, PF-loc's localization accuracy is around 20 m, improved to within 5 m with the spatial verification-based observation model. The global feature-based observation model adjusts the convergence directions of the particles only relying on the overall similarity of scenes, resulting in poor performance in scenes with insufficient and indistinct features. Conversely, the observation model incorporating spatial verification adjusts the convergence directions of the particles relying on the overall similarity of scenes as well as the similarity and positional relationships of the local structures, significantly enhancing localization robustness and accuracy.

\subsection{Necessity of switching the localization mode}\label{subsec_necessity_mode_switch}
\begin{figure*}
\centering
\includegraphics[width=1\linewidth]{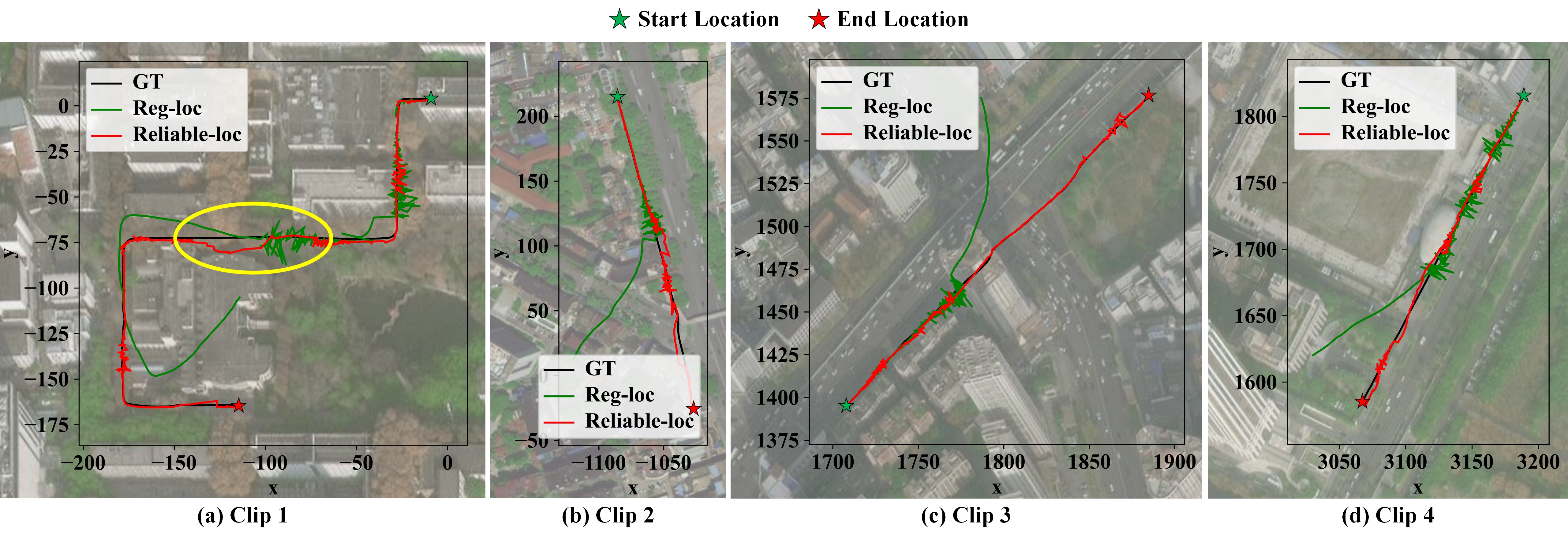}
\caption{Localization trajectories of Reg-loc and Reliable-loc on four typical scenes. The black, green, and red lines are the trajectories of the Ground Truth, Reg-loc, and Reliable-loc, respectively, with the green pentagram as the starting point and the red pentagram as the end point.}\label{fig_necessity_mode_switch}
\end{figure*}

To illustrate the necessity of switching the localization mode, we select representative clips of lengths 750, 500, 500, and 500 frames on the data Info campus, Zhongshan park, Jiefang road 1, and Yanjiang road 1 respectively. We use Reg-loc and Reliable-loc to perform global localization in these clips, the position and yaw angle are known at particle initialization, and the localization trajectories are shown in Fig.\ref{fig_necessity_mode_switch}.

In clip 1, Reg-loc relies on odometry for localization once entering the map's absence area, causing error accumulation and significant deviation from the map, resulting in localization failure. Conversely, Reliable-loc switches to $PF$ mode when the pose uncertainty is high, leveraging particles' exploratory capability to enhance the localization robustness. After traveling about 100 m in clip 2, the scene primarily consists of street trees. Reg-loc heavily relies on odometry for localization, causing pose uncertainty to rapidly rise. However, Reliable-loc enhances localization robustness by switching the localization mode. After traveling about 100 m in clip 3, the scene consists solely of viaducts and flat walls, making it difficult to consistently use $Reg$ mode for localization, while MCL is more robust. After traveling about 100 m in clip 4, the scene contains only some street trees and a huge, smooth ellipsoidal building, with very insufficient features. Reliable-loc effectively enhances localization robustness by switching to $PF$ localization mode. The ablation study in these four clips with insufficient features or incomplete map coverage indicates that monitoring the localization status and adaptively switching the localization mode based on spatial and temporal verifiable cues can significantly improve localization robustness.

\subsection{Distribution of the localization mode}\label{subsec_distribution_loc_mode}
\begin{figure*}[htbp]
\centering
\includegraphics[width=1\linewidth]{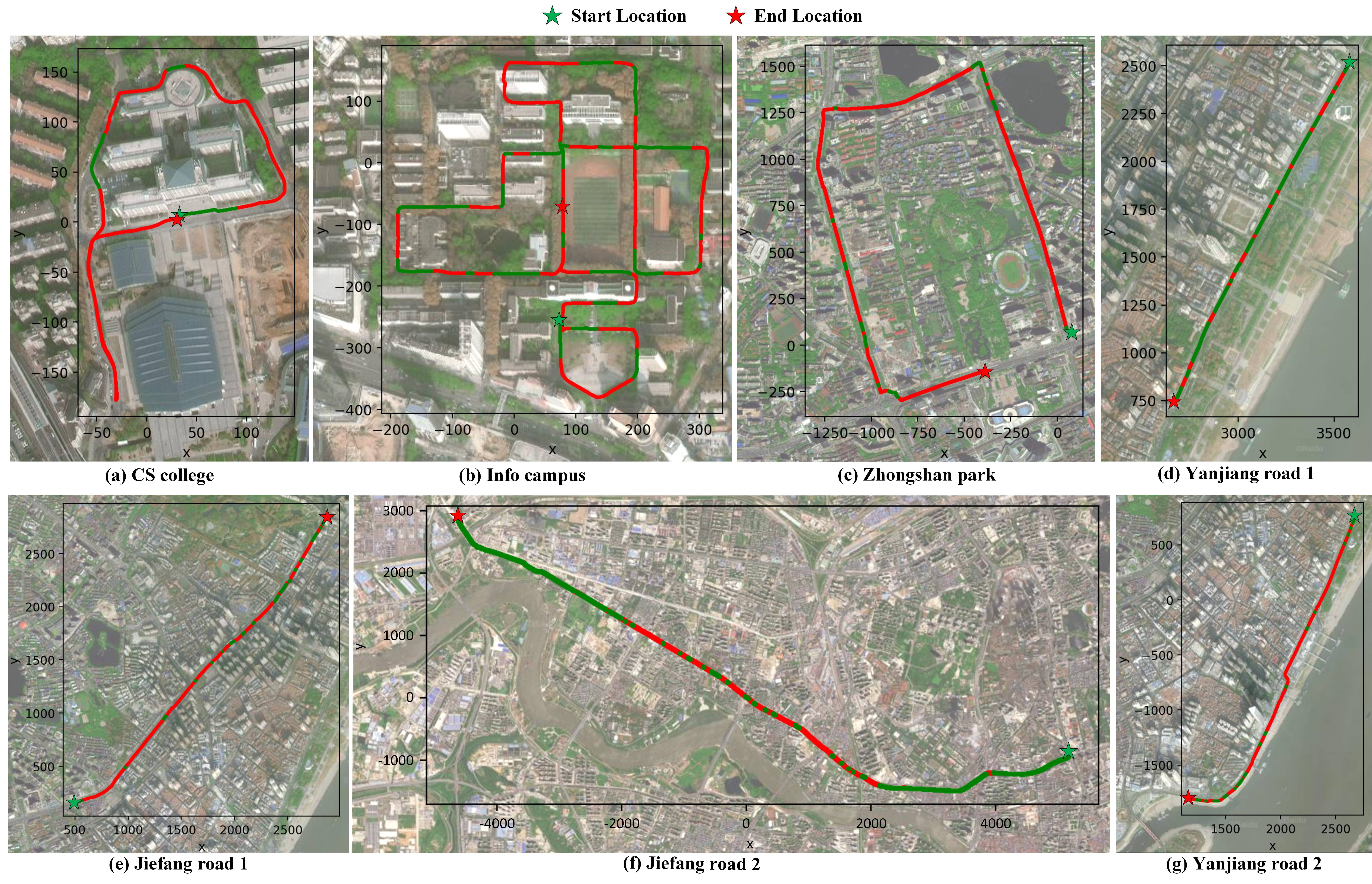}
\caption{Localization mode distribution of Reliable-loc on each data. The green represents the $PF$ localization mode, the red represents the $Reg$ localization mode, the green pentagram is the starting point and the red pentagram is the end point.}\label{fig_distribution_loc_mode}
\end{figure*}

To illustrate the reasonableness of switching the localization mode based on pose uncertainty, we plot the distribution of localization modes and count the proportion of localization modes in each data. Table \ref{table_distribution_loc_mode} shows the proportion of $Reg$ localization modes used by Reliable-loc in each data and Fig.\ref{fig_distribution_loc_mode} depicts the distribution of localization modes employed by Reliable-loc in each data.

As shown in Table \ref{table_distribution_loc_mode}, Reliable-loc predominantly employs the $Reg$ localization mode for most scenes. However, on Yanjiang road 1, the $PF$ localization mode is favored as the helmet-mounted WLS system can only capture flat walls and roadside trees during localization, resulting in feature insufficiency. Fig.\ref{fig_distribution_loc_mode} illustrates that Reliable-loc primarily relies on MCL during the initial stage and in scenes with insufficient features and incomplete map coverage, where pose uncertainty is high. Scenes with insufficient features mainly comprise roadsides with only street trees, proximity to viaducts, and expansive intersections.

\begin{table*}[htbp]
\centering
\fontsize{8}{10}\selectfont
\caption{Localization mode distribution on experimental data.}
\label{table_distribution_loc_mode}
\begin{tblr}{
  width = \linewidth,
  colspec = {Q[156]Q[90]Q[100]Q[125]Q[112]Q[112]Q[119]Q[119]},
  hlines,
}
\textbf{Methods}             & \textbf{CS college} & \textbf{Info campus} & \textbf{Zhongshan park} & \textbf{Jiefang road 1} & \textbf{\textbf{Jiefang road 2}} & \textbf{Yanjiang road 1} & \textbf{Yanjiang road 2} \\
\textbf{Reg Mode Ratio (\%)} & 85.61               & 65.56                & 90.69                   & 83.5                    & 43.34                            & 36.78                    & 87.24                    
\end{tblr}
\end{table*}

\subsection{Parameter analysis}\label{subsec_param_analysis}
The minimum eigenvalue of the Hessian matrix in Subsection \ref{subsec_status_monitor} can reflect the pose uncertainty holistically, which is the most critical parameter for monitoring the localization status, so we analyze the sensitivity of Reliable-loc to the threshold value $\lambda_{thr}$ in detail. We set $\lambda_{thr}$ to 1e-4, 5e-4, 1e-3, 2e-3, 4e-3, 8e-3, and 1.6e-2 respectively, and perform global localization on each data, and the experimental results are shown in Table \ref{table_param_analysis}.

As shown in Table \ref{table_param_analysis}, Reliable-loc exhibits reduced sensitivity to parameters in Info campus, achieving position and yaw accuracies of approximately ±2.5 m and ±4 degrees, respectively. For CS college, Zhongshan park, Jiefang road 1, Jiefang road 2, and Yanjiang road 2, the optimal parameter is 5e-4. For Yanjiang road 1, the optimal parameter is 1e-4. Reliable-loc demonstrates its ability to achieve reliable localization across various parameter settings, maintaining relatively stable performance on the experimental data. Typically, optimal localization performance is achieved when $\lambda_{thr}$ is set to 1e-4, 5e-4, or 1e-3. In practical applications, a setting of 5e-4 for $\lambda_{thr}$ is recommended.

\begin{table*}[htbp]
\centering
\fontsize{7}{10}\selectfont
\caption{Localization accuracy of Reliable-loc with different $\lambda_{thr}$.}
\label{table_param_analysis}
\begin{tblr}{
  width = \linewidth,
  colspec = {Q[113]Q[108]Q[83]Q[92]Q[113]Q[102]Q[102]Q[110]Q[110]},
  cell{2}{1} = {r=2}{},
  cell{4}{1} = {r=2}{},
  cell{6}{1} = {r=2}{},
  cell{8}{1} = {r=2}{},
  cell{10}{1} = {r=2}{},
  cell{12}{1} = {r=2}{},
  cell{14}{1} = {r=2}{},
  hline{1-2,4,6,8,10,12,14,16} = {-}{},
}
\textbf{Param (unit:e-4)} & \textbf{Metrics}     & \textbf{CS college} & \textbf{Info campus} & \textbf{Zhongshan park} & \textbf{Jiefang road 1} & \textbf{\textbf{Jiefang road 2}} & \textbf{Yanjiang road 1} & \textbf{Yanjiang road 2} \\
\textbf{1}                & \textbf{PE (m)↓} & 0.87                & 2.56                 & 1.47                    & \textbf{1.92}           & 5.23                             & \textbf{2.39}            & 1.51                     \\
                          & \textbf{YE (deg)↓}    & 2.17                & 4.19                 & 2.69                    & 2.81                    & 5.23                             & 4.45                     & 3.37                     \\
\textbf{5}                & \textbf{PE (m)↓} & \textbf{0.73}       & 2.41                 & \textbf{1.13}           & \textbf{1.92}           & 4.93                             & 3.60                     & \textbf{1.33}            \\
                          & \textbf{YE (deg)↓}    & \textbf{2.09}       & 4.02                 & \textbf{2.44}           & \textbf{2.64}           & \textbf{4.78}                    & 5.62                     & \textbf{3.22}            \\
\textbf{10}               & \textbf{PE (m)↓} & 0.90                & 2.30                 & 4.14                    & 2.18                    & \textbf{4.77}                    & 2.89                     & 2.18                     \\
                          & \textbf{YE (deg)↓}    & 2.53                & 4.06                 & 3.02                    & 3.24                    & 4.84                             & 4.74                     & 3.16                     \\
\textbf{20}               & \textbf{PE (m)↓} & 0.91                & 2.38                 & 1.43                    & 2.21                    & 5.43                             & 3.06                     & 3.21                     \\
                          & \textbf{YE (deg)↓}    & 2.18                & 4.20                 & 3.09                    & 3.40                    & 4.91                             & \textbf{4.21}            & 4.32                     \\
\textbf{40}               & \textbf{PE (m)↓} & 0.83                & \textbf{2.22}        & 2.27                    & 2.59                    & 5.83                             & 3.18                     & 4.01                     \\
                          & \textbf{YE (deg)↓}    & 2.30                & \textbf{3.66}        & 3.73                    & 4.10                    & 5.17                             & 4.23                     & 4.77                     \\
\textbf{80}               & \textbf{PE (m)↓} & 0.78                & 2.32                 & 2.32                    & 2.75                    & 6.30                             & 3.18                     & 3.95                     \\
                          & \textbf{YE (deg)↓}    & 2.49                & 4.16                 & 3.77                    & 4.06                    & 4.99                             & 4.23                     & 4.55                     \\
\textbf{160}              & \textbf{PE (m)↓} & 1.36                & 2.33                 & 2.94                    & 2.76                    & 6.37                             & 3.18                     & 3.95                     \\
                          & \textbf{YE (deg)↓}    & 2.54                & 4.07                 & 3.99                    & 4.18                    & 5.33                             & 4.23                     & 4.55                     
\end{tblr}
\end{table*}

\subsection{Efficiency analysis}\label{subsec_efficiency_analysis}
In this subsection, we conduct an efficiency analysis of all methods, measuring the average time required for each method to process one submap on the experimental data. Table \ref{table_efficiency_analysis_mean} shows the average time cost of each method. It should be noted that these tables do not include statistics on the time required for extracting features for place recognition.

As shown in Table \ref{table_efficiency_analysis_mean}, Reliable-loc requires an average of 98.04 ms to process one submap. The MCL-based methods exhibit longer total elapsed times, exceeding Reliable-loc by approximately 50 ms. Evidently, Reliable-loc demonstrates superior time efficiency, making it potential for real-time applications.

\begin{table}[ht]
\caption{The average time cost of each method.}\label{table_efficiency_analysis_mean}
\centering
\fontsize{6}{10}\selectfont
\begin{tblr}{
  width = \linewidth,
  colspec = {Q[170]Q[85]Q[145]Q[155]Q[90]Q[170]},
  hlines = {0.08em},
}
\textbf{Methods}            & \textbf{PF-loc} & \textbf{PF-SGV-loc} & \textbf{PF-SGV2-loc} & \textbf{Reg-loc} & \textbf{Reliable-loc (ours)} \\
\textbf{time cost (ms)} & 138.03          & 151.61              & 156.17               & \textbf{45.70}            & \underline{98.04}                 
\end{tblr}
\end{table}

\begin{figure*}[ht]
\centering
\includegraphics[width=1\textwidth]{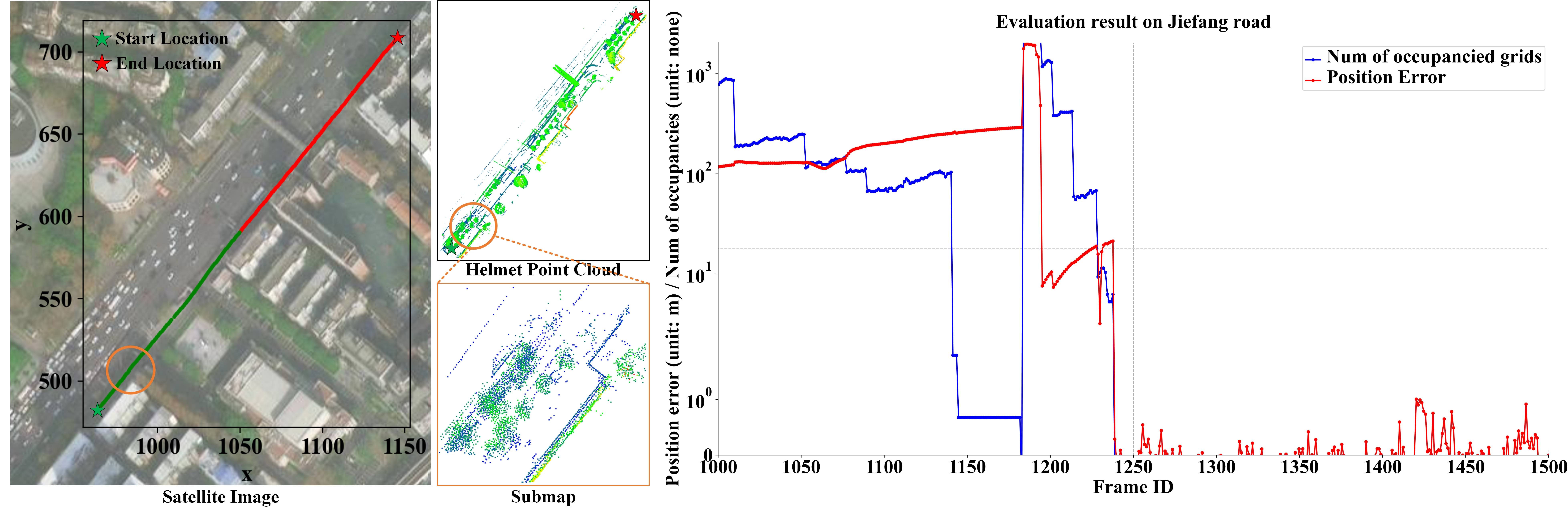}
\caption{Overview of the bad case. The satellite image and trajectory are on the left, with the green line depicting the $PF$ localization mode and the red line indicating the $Reg$ localization mode. The corresponding helmet point cloud is in the middle. Curves of the positioning error and the number of grids occupied by particles over time are on the right, with the red curve depicting the positioning error and the blue curve indicating the number of grids occupied by the particles.}\label{fig_bad_case}
\end{figure*}

\subsection{Deficiencies and future work}\label{subsec_deficiency_future_work}
The above experimental results demonstrate that Reliable-loc can achieve reliable localization in large-scale, complex street scenes by integrating place recognition and MCL, with a position accuracy of up to ±0.73 m and yaw accuracy of up to ±2.09 degrees. Nevertheless, the method's robustness and accuracy rely on the descriptiveness and discrimination of the global and local features from place recognition. When Reliable-loc starts localization in a scene with extremely insufficient features, the particles are prone to converging to the incorrect region due to the poor discrimination of the global features from place recognition, ultimately resulting in localization failure. A corresponding failure case is shown in Fig.\ref{fig_bad_case}, and Fig.\ref{fig_bad_case} shows the satellite image, helmet point cloud, the curve of the positional error, and the number of grids occupied by the particles over time.

As depicted in Fig.\ref{fig_bad_case}, MCL erroneously converges to a location over 100m away from the correct location at the 1153$rd$ frame. Subsequently, the localization is reinitialized at the 1190$th$ frame and converges again at the 1231$th$ frame. However, the system switches to the $Reg$ localization mode at the 1240$th$ frame. Notably, Reliable-loc typically requires over 50 frames to converge and exceeds 1 second per frame. Choosing feature-rich scenes to start localization can alleviate this problem, but significantly limits the practical applicability of such methods.

In the future, we will conduct research in four areas to enhance the robustness and accuracy of the localization systems. Firstly, building on existing place recognition efforts, we will introduce contrastive learning to strengthen the descriptiveness, discrimination, and generalization of features extracted by the network, thereby improving the reliability of the localization systems in challenging scenes such as feature insufficiency \citep{Jing_Self_supervised, Pointcontrast, Wu_Spatiotemporal_Self_supervised}. Secondly, we will explore continuous learning to enable the place recognition model to evolve, overcoming challenges posed by scene switching and sensor changes during long-term localization, and enhancing the system's resilience \citep{Incloud, BioSLAM}. Thirdly, referencing the backend optimization framework in Simultaneous Localization and Mapping, we will fully leverage the information contained in raw data from multiple sensors like IMU and LiDAR to achieve more reliable and accurate localization in a tightly coupled manner \citep{chiang2019seamless, Ye_Tightly_coupled, Pan_Tightly_coupled}. Finally, repetitive or featureless terrains pose significant challenges to the robustness of localization, and It’s difficult to resolve the problem with only point clouds. We therefore consider integrating non-visual sensors such as IMU and UWB to enhance the robustness of localization in repetitive and featureless terrains.

\section{Conclusion}\label{sec_conclusion}
In this paper, we propose a LiDAR-based sequential global localization method that effectively overcomes challenges posed by feature insufficiency and incomplete map coverage by utilizing spatial and temporal verifiable cues, enabling reliable localization in large-scale street scenes. First, the novel MCL incorporating spatial verification adjusts particle weights by leveraging the rich information embedded in local features, enhancing localization robustness in feature-insufficient scenes by avoiding particles converging to erroneous regions. Second, the pose uncertainty-guided reliable global localization framework monitors the localization status and adaptively switches the localization mode, further improving localization robustness in scenes with insufficient features and incomplete map coverage by exploiting the exploratory capability of particles in MCL. We validated the effectiveness of our proposed method on a large-scale heterogeneous point cloud dataset, which comprises high-precision vehicle-mounted MLS point clouds and helmet-mounted WLS point clouds. The experimental results demonstrate that our method achieves a position accuracy of ±2.91 m and a yaw accuracy of ±3.74 degrees in large-scale street scenes, with an average time cost of approximately 98 ms per submap. It exhibits excellent performance in terms of robustness, accuracy, and efficiency. Furthermore, ablation experiments confirm the rationality and effectiveness of the MCL method using spatial verification and the reliable global localization framework guided by pose uncertainty. In the future, we aim to enhance the descriptiveness, discrimination, and generalization of features from place recognition, and to fuse multi-sensor information, such as from IMU, in a tightly coupled manner, to further improve localization performance.

\section*{Declaration of Competing Interest}
The authors declare that they have no known competing financial interests or personal relationships that could have appeared to influence the work reported in this paper.

\section*{Acknowledgments}
This study was jointly supported by the National Natural Science Foundation Project (No.42130105, No.42201477), the National Key Research and Development Program of China (No.2022YFB3904100), the Open Fund of Hubei Luojia Laboratory (No.2201000054).

\appendix
\section{Supplementary for quantitative evaluation}\label{appendix_A}
Table \ref{table_quantitative_eval_ours} shows the localization accuracy of Reliable-loc in different localization modes. As shown in the figure, in $PF$ mode, the position accuracy of Reliable-loc ranges from ±1.90 m to ±7.82 m, and the yaw accuracy ranges from ±3.64 to ±7.86 degrees. In $Reg$ mode, the position accuracy of Reliable-loc ranges from ±0.62 m to ±1.67 m and the yaw accuracy ranges from ±1.95 to ±4.39 degrees. The localization accuracy in $Reg$ mode is much higher than that of in $PF$ mode.

\begin{table*}[htbp]
\centering
\fontsize{7}{10}\selectfont
\caption{Localization success rate and accuracy of Reliable-loc in different localization modes.}
\label{table_quantitative_eval_ours}
\begin{tblr}{
  width = \linewidth,
  colspec = {Q[81]Q[115]Q[87]Q[96]Q[117]Q[106]Q[106]Q[113]Q[113]},
  cell{2}{1} = {r=3}{},
  cell{5}{1} = {r=3}{},
  hline{1-2,5,8} = {-}{},
}
\textbf{Loc Mode} & \textbf{Metrics}      & \textbf{CS college} & \textbf{Info campus} & \textbf{Zhongshan park} & \textbf{Jiefang road 1} & \textbf{\textbf{Jiefang road 2}} & \textbf{Yanjiang road 1} & \textbf{Yanjiang road 2} \\
\textbf{PF}       & \textbf{R@2m5d (\%)↑} & 36.03               & 14.86                & 11.49                   & 25.70                   & 0.60                             & 21.58                    & 9.59                     \\
                  & \textbf{PE (m)↓}      & 1.90 ± 2.00         & 4.48 ± 5.22          & 6.29 ± 35.23            & 3.32 ± 4.13             & 7.82 ± 8.90                      & 3.10 ± 3.66              & 5.19 ± 6.21              \\
                  & \textbf{YE (deg)↓}    & 3.64 ± 4.45         & 5.49 ± 7.66          & 7.86 ± 10.58            & 5.33 ± 6.76             & 5.80 ± 8.20                      & 4.49 ± 5.78              & 6.67 ± 8.46              \\
\textbf{Reg}      & \textbf{R@2m5d (\%)↑} & 87.20               & 71.86                & 87.56                   & 82.00                   & 70.34                            & 61.88                    & 77.68                    \\
                  & \textbf{PE (m)↓}      & 0.62 ± 0.97         & 1.38 ± 1.71          & 0.65 ± 1.52             & 1.67 ± 5.53             & 1.18 ± 1.84                      & 1.22 ± 1.79              & 0.87 ± 1.70              \\
                  & \textbf{YE (deg)↓}    & 1.95 ± 3.45         & 3.29 ± 5.19          & 1.94 ± 3.55             & 2.16 ± 3.87             & 3.45 ± 5.19                      & 4.39 ± 6.39              & 2.80 ± 4.69              
\end{tblr}
\emph{Note: The results in the $PF$ localization mode do not count data before MCL converges.}
\end{table*}

Fig.\ref{fig_quantitative_eval_pva} shows the PVA (position, velocity, and angle) errors of the proposed method on the experimental data. As shown in the figure, Reliable-loc performs well in data of easy and medium difficulty, such as CS college and Zhongshan park. In data of hard difficulty, such as Info campus and Jiefang road 2, the lack of features in the scene, data holes in the map, and changes in LiDAR sensor types have negative impacts on the localization accuracy of the proposed method.

\begin{figure*}
    \centering
    \includegraphics[width=1\linewidth]{figs/Fig4-Quantitative-evaluation-pva.jpg}
    \caption{PVA errors of the proposed method. The black color represents the velocity of the WLS at different time points, while the green, red, and blue colors represent the localization errors of the proposed method at different time points, respectively.}
    \label{fig_quantitative_eval_pva}
\end{figure*}

\section{Supplementary for parameter analysis}\label{appendix_B}
Besides, four 500-frame clips, 0-500 and 500-100 from CS college, 500-1000, 1500-2000 from Jiefang road 1, are used to analyze the impact of the number of particles on localization during initialization and after the convergence of particle filtering. Table \ref{table_param_analysis_num_init_particle} shows the localization accuracy and time cost of Reliable-loc with different numbers of initial particles. As shown in Table \ref{table_param_analysis_num_init_particle}, Reliable-loc performs well in CS college with no less than 500 initial particles and it performs well in Jiefang road 1 with no less than 1000 initial particles. Its time cost rapidly increases with the number of initial particles. Table \ref{table_param_analysis_num_conv_particle} shows the localization accuracy of Reliable-loc with different numbers of particles after the convergence of particle filtering. As shown in Table \ref{table_param_analysis_num_conv_particle}, the number of particles after the convergency of particle filtering has a minor impact on localization performance, and this paper selects a relatively large number of particles to ensure localization robustness.

\begin{table*}[htbp]
\centering
\fontsize{7}{10}\selectfont
\caption{Localization accuracy and time cost of Reliable-loc with different numbers of initial particles.}
\label{table_param_analysis_num_init_particle}
\begin{tblr}{
  width = \linewidth,
  colspec = {Q[106]Q[154]Q[127]Q[127]Q[106]Q[106]Q[106]Q[106]},
  cell{2}{1} = {r=3}{},
  cell{5}{1} = {r=3}{},
  cell{8}{1} = {r=3}{},
  cell{11}{1} = {r=3}{},
  hline{1-2,5,8,11,14} = {-}{},
}
\textbf{Data clips} & {\textbf{Num of}\\\textbf{initial particles}} & \textbf{100} & \textbf{500} & \textbf{1000} & \textbf{2500} & \textbf{5000} & \textbf{10000} \\
\textbf{1}          & \textbf{PE (m)↓}                                        & 44.41 ± 48.16  & 8.15 ± 115.84  & 0.84 ± 6.34     & 0.61 ± 1.55     & 0.79 ± 2.28     & 0.57 ± 0.95      \\
                    & \textbf{YE (deg)↓}                                      & 22.98 ± 28.01  & 2.93 ± 13.18   & 2.51 ± 7.48     & 2.38 ± 8.72     & 2.45 ± 4.63     & 2.07 ± 3.68      \\
                    & \textbf{time cost (ms)↓}                                & 132.57       & 117.4        & 123.03        & 162.28        & 369.44        & 702.1          \\
\textbf{2}          & \textbf{PE (m)↓}                                        & 46.07 ± 74.63  & 19.38 ± 36.84  & 0.52 ± 1.00     & 0.65 ± 1.35     & 0.64 ± 1.19     & 0.48 ± 0.78      \\
                    & \textbf{YE (deg)↓}                                      & 35.17 ± 56.20  & 16.85 ± 32.95  & 2.04 ± 3.53     & 2.60 ± 6.48     & 2.36 ± 4.48     & 1.85 ± 3.18      \\
                    & \textbf{time cost (ms)↓}                                & 112.44       & 158.93       & 125.39        & 172.32        & 271.7         & 659.73         \\
\textbf{3}          & \textbf{PE (m)↓}                                        & 1.81 ± 3.37    & 0.85 ± 1.58    & 1.02 ± 1.54     & 0.74 ± 1.19     & 1.33 ± 2.00     & 1.10 ± 1.66      \\
                    & \textbf{YE (deg)↓}                                      & 3.83 ± 7.82    & 2.33 ± 4.04    & 2.58 ± 4.48     & 2.37 ± 4.11     & 4.33 ± 6.93     & 3.35 ± 5.48      \\
                    & \textbf{time cost (ms)↓}                                & 112.59       & 118          & 133.38        & 171.72        & 440.26        & 474.28         \\
\textbf{4}          & \textbf{PE (m)↓}                                        & 46.73 ± 48.44  & 1.13 ± 1.66    & 1.50 ± 3.21     & 1.35 ± 2.10     & 1.85 ± 3.46     & 0.99 ± 2.38      \\
                    & \textbf{YE (deg)↓}                                      & 23.14 ± 32.31  & 2.60 ± 4.11    & 3.38 ± 8.68     & 2.58 ± 4.06     & 4.05 ± 8.80     & 2.71 ± 5.70      \\
                    & \textbf{time cost (ms)↓}                                & 129.41       & 152.94       & 149.59        & 194.51        & 220.04        & 459.72         
\end{tblr}
\end{table*}

\begin{table*}[htbp]
\centering
\fontsize{7}{10}\selectfont
\caption{Localization accuracy of Reliable-loc with different numbers of particles after the convergence of particle filtering.}
\label{table_param_analysis_num_conv_particle}
\begin{tblr}{
  width = \linewidth,
  colspec = {Q[98]Q[171]Q[113]Q[113]Q[113]Q[113]Q[104]Q[104]},
  cell{2}{1} = {r=2}{},
  cell{4}{1} = {r=2}{},
  cell{6}{1} = {r=2}{},
  cell{8}{1} = {r=2}{},
  hline{1-2,4,6,8,10} = {-}{},
}
\textbf{Data clips} & {\textbf{Num of particles}\\\textbf{after convergence}} & \textbf{25}  & \textbf{50}  & \textbf{100} & \textbf{200} & \textbf{400} & \textbf{800} \\
\textbf{1}          & \textbf{PE (m)↓}                                                  & 0.86 ± 2.38  & 0.76 ± 2.32  & 0.77 ± 2.35  & 0.83 ± 2.41  & 0.87 ± 2.41  & 0.90 ± 2.58  \\
                    & \textbf{YE (deg)↓}                                                & 2.52 ± 4.84  & 2.18 ± 4.39  & 2.27 ± 4.54  & 2.14 ± 4.16  & 2.53 ± 5.07  & 2.28 ± 4.32  \\
\textbf{2}          & \textbf{PE (m)↓}                                                  & 0.50 ± 0.84  & 0.59 ± 1.06  & 0.54 ± 0.95  & 0.58 ± 1.10  & 0.66 ± 1.22  & 0.61 ± 1.14  \\
                    & \textbf{YE (deg)↓}                                                & 1.78 ± 3.24  & 2.08 ± 3.89  & 1.87 ± 3.53  & 2.12 ± 3.99  & 2.36 ± 4.71  & 2.01 ± 3.70  \\
\textbf{3}          & \textbf{PE (m)↓}                                                  & 1.59 ± 2.47  & 0.99 ± 1.44  & 0.93 ± 1.39  & 1.27 ± 2.21  & 1.18 ± 1.70  & 1.18 ± 1.76  \\
                    & \textbf{YE (deg)↓}                                                & 3.62 ± 6.35  & 3.25 ± 5.50  & 3.13 ± 4.95  & 3.95 ± 6.20  & 3.49 ± 5.56  & 3.63 ± 5.57  \\
\textbf{4}          & \textbf{PE (m)↓}                                                  & 1.54 ± 3.65  & 1.28 ± 3.50  & 2.27 ± 4.16  & 1.85 ± 3.94  & 1.35 ± 2.94  & 2.14 ± 3.74  \\
                    & \textbf{YE (deg)↓}                                                & 5.47 ± 16.68 & 5.03 ± 16.03 & 5.46 ± 15.91 & 5.33 ± 15.86 & 3.79 ± 8.47  & 3.93 ± 7.13  
\end{tblr}
\end{table*}

\section{Localization robustness to partial observations, dynamic objects, and content changes}\label{appendix_C}
\begin{figure*}[htbp]
    \centering
    \includegraphics[width=1\linewidth]{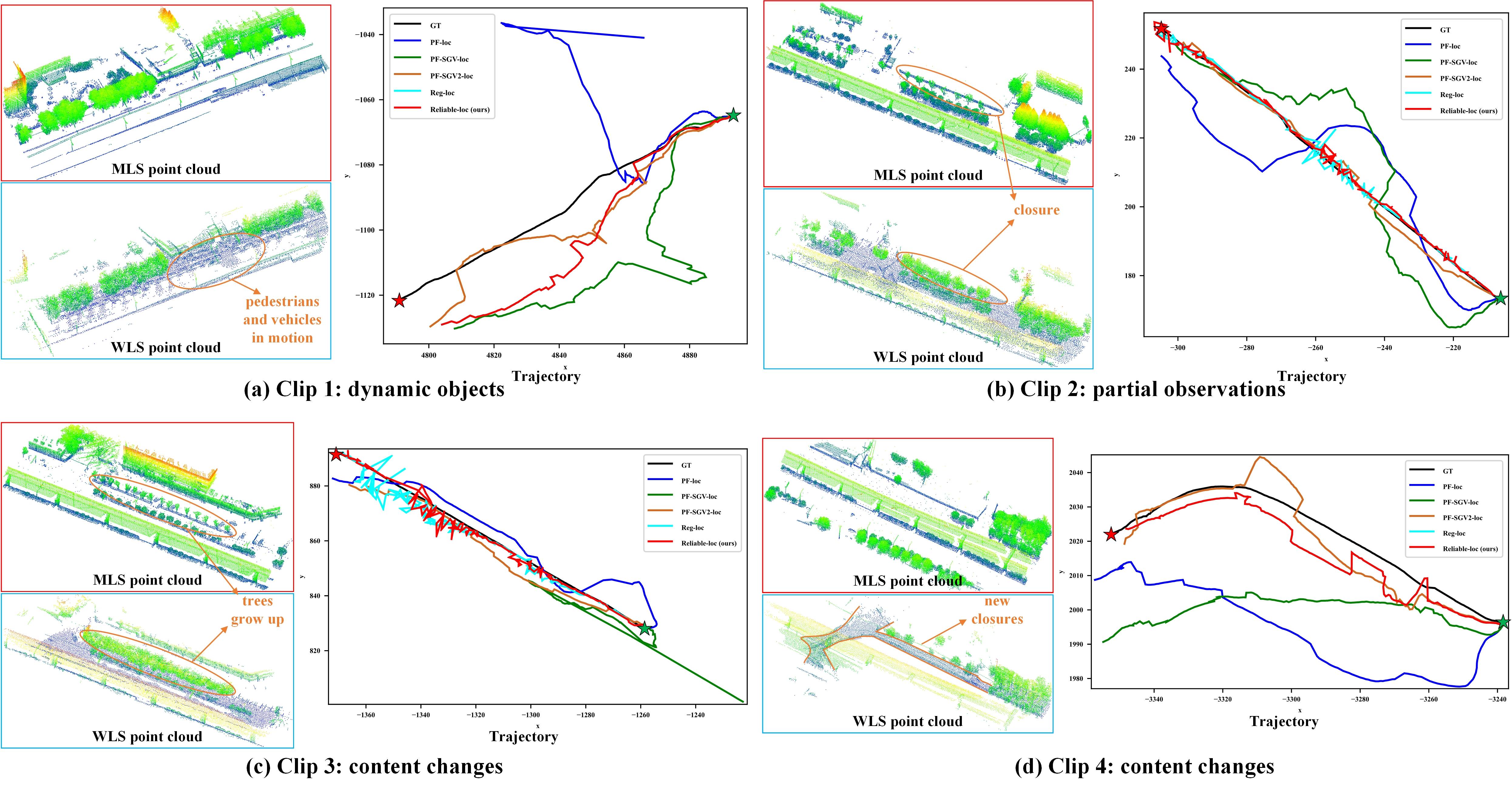}
    \caption{Localization trajectories in the four clips when the initial position is known and the yaw is unknown. The black, blue, green, orange, cyan, and red lines are the trajectories of the Ground Truth, PF-loc, PF-SGV-loc, PF-SGV2-loc, Reg-loc, and Reliable-loc, respectively, with the green pentagram as the starting point and the red pentagram as the end point.}
    \label{fig_cases_jiefang_road2}
\end{figure*}

Although partial observations, dynamic objects, and content changes are not the primary focus of this paper, we still select four 200-frame clips from Jiefang road 2 and conduct additional experiments refer to section \ref{subsec_effectiveness_obs_model_gv} to demonstrate the robustness of our method against these challenges. As shown in Fig.\ref{fig_cases_jiefang_road2}, the localization performance of PF-loc, PF-SGV-loc, PF-SGV2-loc, and Reliable-loc is improved sequentially on all four clips. In clip 1, lots of pedestrians and vehicles pose challenges to localization. In clip 2, occlusions from the long closure limits the MLS and WLS systems to capturing only very limited scene content. In clip 3 and 4, due to a time difference of over 4 years between the MLS and WLS point cloud acquisitions, trees in some scenes have grown significantly, and construction closures have been newly added in other scenes. Owing to the proposed spatial verification-based MCL and localization status monitoring mechanism, Reliable-loc performs well in all four clips.

\bibliographystyle{elsarticle-harv} 
\bibliography{cas-refs}

\begin{thebibliography}{63}
\expandafter\ifx\csname natexlab\endcsname\relax\def\natexlab#1{#1}\fi
\providecommand{\url}[1]{\texttt{#1}}
\providecommand{\href}[2]{#2}
\providecommand{\path}[1]{#1}
\providecommand{\DOIprefix}{doi:}
\providecommand{\ArXivprefix}{arXiv:}
\providecommand{\URLprefix}{URL: }
\providecommand{\Pubmedprefix}{pmid:}
\providecommand{\doi}[1]{\href{http://dx.doi.org/#1}{\path{#1}}}
\providecommand{\Pubmed}[1]{\href{pmid:#1}{\path{#1}}}
\providecommand{\bibinfo}[2]{#2}
\ifx\xfnm\relax \def\xfnm[#1]{\unskip,\space#1}\fi
\bibitem[{Akai et~al.(2020)Akai, Hirayama and Murase}]{Akai_Hybrid}
\bibinfo{author}{Akai, N.}, \bibinfo{author}{Hirayama, T.},
  \bibinfo{author}{Murase, H.}, \bibinfo{year}{2020}.
\newblock \bibinfo{title}{Hybrid localization using model-and learning-based
  methods: Fusion of monte carlo and e2e localizations via importance
  sampling}, in: \bibinfo{booktitle}{2020 IEEE International Conference on
  Robotics and Automation (ICRA)}, \bibinfo{organization}{IEEE}. pp.
  \bibinfo{pages}{6469--6475}.
\bibitem[{Almqvist et~al.(2018)Almqvist, Magnusson, Kucner and
  Lilienthal}]{Almqvist_Learning}
\bibinfo{author}{Almqvist, H.}, \bibinfo{author}{Magnusson, M.},
  \bibinfo{author}{Kucner, T.P.}, \bibinfo{author}{Lilienthal, A.},
  \bibinfo{year}{2018}.
\newblock \bibinfo{title}{Learning to detect misaligned point clouds}.
\newblock \bibinfo{journal}{Journal of Field Robotics} \bibinfo{volume}{35},
  \bibinfo{pages}{662--677}.
\bibitem[{Alsayed et~al.(2017)Alsayed, Bresson, Verroust-Blondet and
  Nashashibi}]{Alsayed_Failure}
\bibinfo{author}{Alsayed, Z.}, \bibinfo{author}{Bresson, G.},
  \bibinfo{author}{Verroust-Blondet, A.}, \bibinfo{author}{Nashashibi, F.},
  \bibinfo{year}{2017}.
\newblock \bibinfo{title}{Failure detection for laser-based slam in urban and
  peri-urban environments}, in: \bibinfo{booktitle}{2017 IEEE 20th
  International Conference on Intelligent Transportation Systems (ITSC)},
  \bibinfo{organization}{IEEE}. pp. \bibinfo{pages}{1--7}.
\bibitem[{Arandjelovic et~al.(2016)Arandjelovic, Gronat, Torii, Pajdla and
  Sivic}]{NetVLAD}
\bibinfo{author}{Arandjelovic, R.}, \bibinfo{author}{Gronat, P.},
  \bibinfo{author}{Torii, A.}, \bibinfo{author}{Pajdla, T.},
  \bibinfo{author}{Sivic, J.}, \bibinfo{year}{2016}.
\newblock \bibinfo{title}{Netvlad: Cnn architecture for weakly supervised place
  recognition}, in: \bibinfo{booktitle}{Proceedings of the IEEE conference on
  computer vision and pattern recognition}, pp. \bibinfo{pages}{5297--5307}.
\bibitem[{Arthur et~al.(2007)Arthur, Vassilvitskii et~al.}]{k_means_plus}
\bibinfo{author}{Arthur, D.}, \bibinfo{author}{Vassilvitskii, S.}, et~al.,
  \bibinfo{year}{2007}.
\newblock \bibinfo{title}{k-means++: The advantages of careful seeding}, in:
  \bibinfo{booktitle}{Soda}, pp. \bibinfo{pages}{1027--1035}.
\bibitem[{Baglietto et~al.(2011)Baglietto, Sgorbissa, Verda and
  Zaccaria}]{Baglietto_Human}
\bibinfo{author}{Baglietto, M.}, \bibinfo{author}{Sgorbissa, A.},
  \bibinfo{author}{Verda, D.}, \bibinfo{author}{Zaccaria, R.},
  \bibinfo{year}{2011}.
\newblock \bibinfo{title}{Human navigation and mapping with a 6dof imu and a
  laser scanner}.
\newblock \bibinfo{journal}{Robotics Autonomous Systems} \bibinfo{volume}{59},
  \bibinfo{pages}{1060--1069}.
\bibitem[{Besl and McKay(1992)}]{ICP}
\bibinfo{author}{Besl, P.J.}, \bibinfo{author}{McKay, N.D.},
  \bibinfo{year}{1992}.
\newblock \bibinfo{title}{Method for registration of 3-d shapes}, in:
  \bibinfo{booktitle}{Sensor fusion IV: control paradigms and data structures},
  \bibinfo{organization}{Spie}. pp. \bibinfo{pages}{586--606}.
\bibitem[{Canedo-Rodr{\'\i}guez et~al.(2016)Canedo-Rodr{\'\i}guez,
  Alvarez-Santos, Regueiro, Iglesias, Barro and Presedo}]{canedo2016particle}
\bibinfo{author}{Canedo-Rodr{\'\i}guez, A.}, \bibinfo{author}{Alvarez-Santos,
  V.}, \bibinfo{author}{Regueiro, C.V.}, \bibinfo{author}{Iglesias, R.},
  \bibinfo{author}{Barro, S.}, \bibinfo{author}{Presedo, J.},
  \bibinfo{year}{2016}.
\newblock \bibinfo{title}{Particle filter robot localisation through robust
  fusion of laser, wifi, compass, and a network of external cameras}.
\newblock \bibinfo{journal}{Information Fusion} \bibinfo{volume}{27},
  \bibinfo{pages}{170--188}.
\bibitem[{Cattaneo et~al.(2022)Cattaneo, Vaghi and Valada}]{Lcdnet}
\bibinfo{author}{Cattaneo, D.}, \bibinfo{author}{Vaghi, M.},
  \bibinfo{author}{Valada, A.}, \bibinfo{year}{2022}.
\newblock \bibinfo{title}{Lcdnet: Deep loop closure detection and point cloud
  registration for lidar slam}.
\newblock \bibinfo{journal}{IEEE Transactions on Robotics}
  \bibinfo{volume}{38}, \bibinfo{pages}{2074--2093}.
\bibitem[{Chen et~al.(2021a)Chen, Yin, Jiao, Dissanayake, Wang and
  Xiong}]{DSOM}
\bibinfo{author}{Chen, R.}, \bibinfo{author}{Yin, H.}, \bibinfo{author}{Jiao,
  Y.}, \bibinfo{author}{Dissanayake, G.}, \bibinfo{author}{Wang, Y.},
  \bibinfo{author}{Xiong, R.}, \bibinfo{year}{2021}a.
\newblock \bibinfo{title}{Deep samplable observation model for global
  localization and kidnapping}.
\newblock \bibinfo{journal}{IEEE Robotics Automation Letters}
  \bibinfo{volume}{6}, \bibinfo{pages}{2296--2303}.
\bibitem[{Chen et~al.(2020)Chen, L{\"a}be, Nardi, Behley and
  Stachniss}]{overlap_loc}
\bibinfo{author}{Chen, X.}, \bibinfo{author}{L{\"a}be, T.},
  \bibinfo{author}{Nardi, L.}, \bibinfo{author}{Behley, J.},
  \bibinfo{author}{Stachniss, C.}, \bibinfo{year}{2020}.
\newblock \bibinfo{title}{Learning an overlap-based observation model for 3d
  lidar localization}, in: \bibinfo{booktitle}{2020 IEEE/RSJ International
  Conference on Intelligent Robots and Systems (IROS)},
  \bibinfo{organization}{IEEE}. pp. \bibinfo{pages}{4602--4608}.
\bibitem[{Chen et~al.(2021b)Chen, Läbe, Milioto, Röhling, Vysotska, Haag,
  Behley and Stachniss}]{OverlapNet}
\bibinfo{author}{Chen, X.}, \bibinfo{author}{Läbe, T.},
  \bibinfo{author}{Milioto, A.}, \bibinfo{author}{Röhling, T.},
  \bibinfo{author}{Vysotska, O.}, \bibinfo{author}{Haag, A.},
  \bibinfo{author}{Behley, J.}, \bibinfo{author}{Stachniss, C.},
  \bibinfo{year}{2021}b.
\newblock \bibinfo{title}{Overlapnet: Loop closing for lidar-based slam}.
\newblock \bibinfo{journal}{arXiv} .
\bibitem[{Chi et~al.(2022)Chi, Kim, Liu, Thedja, Seo and Lee}]{Chi_Rebar}
\bibinfo{author}{Chi, H.L.}, \bibinfo{author}{Kim, M.K.}, \bibinfo{author}{Liu,
  K.Z.}, \bibinfo{author}{Thedja, J.}, \bibinfo{author}{Seo, J.},
  \bibinfo{author}{Lee, D.E.}, \bibinfo{year}{2022}.
\newblock \bibinfo{title}{Rebar inspection integrating augmented reality and
  laser scanning}.
\newblock \bibinfo{journal}{Automation in Construction} \bibinfo{volume}{136},
  \bibinfo{pages}{104183}.
\bibitem[{Chiang et~al.(2019)Chiang, Tsai, Chang, Joly and
  Ei-Sheimy}]{chiang2019seamless}
\bibinfo{author}{Chiang, K.W.}, \bibinfo{author}{Tsai, G.J.},
  \bibinfo{author}{Chang, H.}, \bibinfo{author}{Joly, C.},
  \bibinfo{author}{Ei-Sheimy, N.}, \bibinfo{year}{2019}.
\newblock \bibinfo{title}{Seamless navigation and mapping using an
  ins/gnss/grid-based slam semi-tightly coupled integration scheme}.
\newblock \bibinfo{journal}{Information Fusion} \bibinfo{volume}{50},
  \bibinfo{pages}{181--196}.
\bibitem[{Cummins and Newman(2008)}]{FAB_MAP}
\bibinfo{author}{Cummins, M.}, \bibinfo{author}{Newman, P.},
  \bibinfo{year}{2008}.
\newblock \bibinfo{title}{Fab-map: Probabilistic localization and mapping in
  the space of appearance}.
\newblock \bibinfo{journal}{The International Journal of Robotics Research}
  \bibinfo{volume}{27}, \bibinfo{pages}{647--665}.
\bibitem[{Du et~al.(2020)Du, Wang and Cremers}]{Dh3d}
\bibinfo{author}{Du, J.}, \bibinfo{author}{Wang, R.}, \bibinfo{author}{Cremers,
  D.}, \bibinfo{year}{2020}.
\newblock \bibinfo{title}{Dh3d: Deep hierarchical 3d descriptors for robust
  large-scale 6dof relocalization}, in: \bibinfo{booktitle}{Computer
  Vision--ECCV 2020: 16th European Conference, Glasgow, UK, August 23--28,
  2020, Proceedings, Part IV 16}, \bibinfo{organization}{Springer}. pp.
  \bibinfo{pages}{744--762}.
\bibitem[{Fujii et~al.(2015)Fujii, Tanaka, Yabushita, Mori and
  Odashima}]{Fujii_Detection}
\bibinfo{author}{Fujii, A.}, \bibinfo{author}{Tanaka, M.},
  \bibinfo{author}{Yabushita, H.}, \bibinfo{author}{Mori, T.},
  \bibinfo{author}{Odashima, T.}, \bibinfo{year}{2015}.
\newblock \bibinfo{title}{Detection of localization failure using logistic
  regression}, in: \bibinfo{booktitle}{2015 IEEE/RSJ International Conference
  on Intelligent Robots and Systems (IROS)}, \bibinfo{organization}{IEEE}. pp.
  \bibinfo{pages}{4313--4318}.
\bibitem[{Golub and Reinsch(1971)}]{SVD}
\bibinfo{author}{Golub, G.H.}, \bibinfo{author}{Reinsch, C.},
  \bibinfo{year}{1971}.
\newblock \bibinfo{title}{Singular value decomposition and least squares
  solutions}, in: \bibinfo{booktitle}{Handbook for Automatic Computation:
  Volume II: Linear Algebra}. \bibinfo{publisher}{Springer}, pp.
  \bibinfo{pages}{134--151}.
\bibitem[{Goran(2010)}]{Radosevic_Laser_scanning}
\bibinfo{author}{Goran, R.}, \bibinfo{year}{2010}.
\newblock \bibinfo{title}{Laser scanning versus photogrammetry combined with
  manual post-modeling in stecak digitization}, in: \bibinfo{booktitle}{Proc.
  14th Central European Seminar on Computer Graphics},
  \bibinfo{organization}{Citeseer}.
\bibitem[{Guan et~al.(2019)Guan, Ristic, Wang and Palmer}]{guan2019kld}
\bibinfo{author}{Guan, R.P.}, \bibinfo{author}{Ristic, B.},
  \bibinfo{author}{Wang, L.}, \bibinfo{author}{Palmer, J.L.},
  \bibinfo{year}{2019}.
\newblock \bibinfo{title}{Kld sampling with gmapping proposal for monte carlo
  localization of mobile robots}.
\newblock \bibinfo{journal}{Information Fusion} \bibinfo{volume}{49},
  \bibinfo{pages}{79--88}.
\bibitem[{Han et~al.(2024)Han, Liu, Zhou, Tan, Dong and Yang}]{han2024whu}
\bibinfo{author}{Han, X.}, \bibinfo{author}{Liu, C.}, \bibinfo{author}{Zhou,
  Y.}, \bibinfo{author}{Tan, K.}, \bibinfo{author}{Dong, Z.},
  \bibinfo{author}{Yang, B.}, \bibinfo{year}{2024}.
\newblock \bibinfo{title}{Whu-urban3d: An urban scene lidar point cloud dataset
  for semantic instance segmentation}.
\newblock \bibinfo{journal}{ISPRS Journal of Photogrammetry and Remote Sensing}
  \bibinfo{volume}{209}, \bibinfo{pages}{500--513}.
\bibitem[{Jing and Tian(2020)}]{Jing_Self_supervised}
\bibinfo{author}{Jing, L.}, \bibinfo{author}{Tian, Y.}, \bibinfo{year}{2020}.
\newblock \bibinfo{title}{Self-supervised visual feature learning with deep
  neural networks: A survey}.
\newblock \bibinfo{journal}{IEEE transactions on pattern analysis machine
  intelligence} \bibinfo{volume}{43}, \bibinfo{pages}{4037--4058}.
\bibitem[{Kachurka et~al.(2021)Kachurka, Rault, Muñoz, Roussel, Bonardi,
  Didier, Hadj-Abdelkader, Bouchafa, Alliez and Robin}]{WeCo_SLAM}
\bibinfo{author}{Kachurka, V.}, \bibinfo{author}{Rault, B.},
  \bibinfo{author}{Muñoz, F.I.I.}, \bibinfo{author}{Roussel, D.},
  \bibinfo{author}{Bonardi, F.}, \bibinfo{author}{Didier, J.Y.},
  \bibinfo{author}{Hadj-Abdelkader, H.}, \bibinfo{author}{Bouchafa, S.},
  \bibinfo{author}{Alliez, P.}, \bibinfo{author}{Robin, M.},
  \bibinfo{year}{2021}.
\newblock \bibinfo{title}{Weco-slam: Wearable cooperative slam system for
  real-time indoor localization under challenging conditions}.
\newblock \bibinfo{journal}{IEEE Sensors Journal} \bibinfo{volume}{22},
  \bibinfo{pages}{5122--5132}.
\bibitem[{Kirsch et~al.(2022)Kirsch, G{\"u}nter and
  K{\"o}nig}]{Kirsch_Predicting}
\bibinfo{author}{Kirsch, A.}, \bibinfo{author}{G{\"u}nter, A.},
  \bibinfo{author}{K{\"o}nig, M.}, \bibinfo{year}{2022}.
\newblock \bibinfo{title}{Predicting alignability of point cloud pairs for
  point cloud registration using features}, in: \bibinfo{booktitle}{2022 12th
  International Conference on Pattern Recognition Systems (ICPRS)},
  \bibinfo{organization}{IEEE}. pp. \bibinfo{pages}{1--6}.
\bibitem[{Knights et~al.(2022)Knights, Moghadam, Ramezani, Sridharan and
  Fookes}]{Incloud}
\bibinfo{author}{Knights, J.}, \bibinfo{author}{Moghadam, P.},
  \bibinfo{author}{Ramezani, M.}, \bibinfo{author}{Sridharan, S.},
  \bibinfo{author}{Fookes, C.}, \bibinfo{year}{2022}.
\newblock \bibinfo{title}{Incloud: Incremental learning for point cloud place
  recognition}, in: \bibinfo{booktitle}{2022 IEEE/RSJ International Conference
  on Intelligent Robots and Systems (IROS)}, \bibinfo{organization}{IEEE}. pp.
  \bibinfo{pages}{8559--8566}.
\bibitem[{Komorowski(2021)}]{Minkloc3d}
\bibinfo{author}{Komorowski, J.}, \bibinfo{year}{2021}.
\newblock \bibinfo{title}{Minkloc3d: Point cloud based large-scale place
  recognition}, in: \bibinfo{booktitle}{Proceedings of the IEEE/CVF Winter
  Conference on Applications of Computer Vision}, pp.
  \bibinfo{pages}{1790--1799}.
\bibitem[{Komorowski et~al.(2021)Komorowski, Wysoczanska and Trzcinski}]{Egonn}
\bibinfo{author}{Komorowski, J.}, \bibinfo{author}{Wysoczanska, M.},
  \bibinfo{author}{Trzcinski, T.}, \bibinfo{year}{2021}.
\newblock \bibinfo{title}{Egonn: Egocentric neural network for point cloud
  based 6dof relocalization at the city scale}.
\newblock \bibinfo{journal}{IEEE Robotics Automation Letters}
  \bibinfo{volume}{7}, \bibinfo{pages}{722--729}.
\bibitem[{Leordeanu and Hebert(2005)}]{Spectral_matching}
\bibinfo{author}{Leordeanu, M.}, \bibinfo{author}{Hebert, M.},
  \bibinfo{year}{2005}.
\newblock \bibinfo{title}{A spectral technique for correspondence problems
  using pairwise constraints}, in: \bibinfo{booktitle}{Tenth IEEE International
  Conference on Computer Vision (ICCV'05) Volume 1},
  \bibinfo{organization}{IEEE}. pp. \bibinfo{pages}{1482--1489}.
\bibitem[{Li et~al.(2023)Li, Wu, Yang, Zou, Yang, Zhao and Dong}]{WHU_Helmet}
\bibinfo{author}{Li, J.}, \bibinfo{author}{Wu, W.}, \bibinfo{author}{Yang, B.},
  \bibinfo{author}{Zou, X.}, \bibinfo{author}{Yang, Y.}, \bibinfo{author}{Zhao,
  X.}, \bibinfo{author}{Dong, Z.}, \bibinfo{year}{2023}.
\newblock \bibinfo{title}{Whu-helmet: A helmet-based multi-sensor slam dataset
  for the evaluation of real-time 3d mapping in large-scale gnss-denied
  environments}.
\newblock \bibinfo{journal}{IEEE Transactions on Geoscience and Remote Sensing}
  .
\bibitem[{Li et~al.(2024)Li, Yuan, Cao, Nguyen, Cao and Xie}]{li2024hcto}
\bibinfo{author}{Li, J.}, \bibinfo{author}{Yuan, S.}, \bibinfo{author}{Cao,
  M.}, \bibinfo{author}{Nguyen, T.M.}, \bibinfo{author}{Cao, K.},
  \bibinfo{author}{Xie, L.}, \bibinfo{year}{2024}.
\newblock \bibinfo{title}{Hcto: Optimality-aware lidar inertial odometry with
  hybrid continuous time optimization for compact wearable mapping system}.
\newblock \bibinfo{journal}{arXiv preprint arXiv:2403.14173} .
\bibitem[{Li et~al.(2021)Li, Yang, Weng and Wang}]{Li_Robust_Loc}
\bibinfo{author}{Li, L.}, \bibinfo{author}{Yang, M.}, \bibinfo{author}{Weng,
  L.}, \bibinfo{author}{Wang, C.}, \bibinfo{year}{2021}.
\newblock \bibinfo{title}{Robust localization for intelligent vehicles based on
  pole-like features using the point cloud}.
\newblock \bibinfo{journal}{IEEE Transactions on Automation Science
  Engineering} \bibinfo{volume}{PP}, \bibinfo{pages}{1--14}.
\bibitem[{Liu et~al.(2019a)Liu, Suo, Zhou, Xu, Wei, Chen, Wang, Liang and
  Liu}]{Seqlpd}
\bibinfo{author}{Liu, Z.}, \bibinfo{author}{Suo, C.}, \bibinfo{author}{Zhou,
  S.}, \bibinfo{author}{Xu, F.}, \bibinfo{author}{Wei, H.},
  \bibinfo{author}{Chen, W.}, \bibinfo{author}{Wang, H.},
  \bibinfo{author}{Liang, X.}, \bibinfo{author}{Liu, Y.H.},
  \bibinfo{year}{2019}a.
\newblock \bibinfo{title}{Seqlpd: Sequence matching enhanced loop-closure
  detection based on large-scale point cloud description for self-driving
  vehicles}, in: \bibinfo{booktitle}{2019 IEEE/RSJ International Conference on
  Intelligent Robots and Systems (IROS)}, \bibinfo{organization}{IEEE}. pp.
  \bibinfo{pages}{1218--1223}.
\bibitem[{Liu et~al.(2019b)Liu, Zhou, Suo, Yin, Chen, Wang, Li and
  Liu}]{Lpd_net}
\bibinfo{author}{Liu, Z.}, \bibinfo{author}{Zhou, S.}, \bibinfo{author}{Suo,
  C.}, \bibinfo{author}{Yin, P.}, \bibinfo{author}{Chen, W.},
  \bibinfo{author}{Wang, H.}, \bibinfo{author}{Li, H.}, \bibinfo{author}{Liu,
  Y.H.}, \bibinfo{year}{2019}b.
\newblock \bibinfo{title}{Lpd-net: 3d point cloud learning for large-scale
  place recognition and environment analysis}, in:
  \bibinfo{booktitle}{Proceedings of the IEEE/CVF international conference on
  computer vision}, pp. \bibinfo{pages}{2831--2840}.
\bibitem[{Ma et~al.(2022)Ma, Chen, Xu and Xiong}]{SeqOT}
\bibinfo{author}{Ma, J.}, \bibinfo{author}{Chen, X.}, \bibinfo{author}{Xu, J.},
  \bibinfo{author}{Xiong, G.}, \bibinfo{year}{2022}.
\newblock \bibinfo{title}{Seqot: A spatial-temporal transformer network for
  place recognition using sequential lidar data}.
\newblock \bibinfo{journal}{IEEE Transactions on Industrial Electronics} .
\bibitem[{Mi et~al.(2021)Mi, Yang, Dong, Chen and Gu}]{mi2021automated}
\bibinfo{author}{Mi, X.}, \bibinfo{author}{Yang, B.}, \bibinfo{author}{Dong,
  Z.}, \bibinfo{author}{Chen, C.}, \bibinfo{author}{Gu, J.},
  \bibinfo{year}{2021}.
\newblock \bibinfo{title}{Automated 3d road boundary extraction and
  vectorization using mls point clouds}.
\newblock \bibinfo{journal}{IEEE Transactions on Intelligent Transportation
  Systems} \bibinfo{volume}{23}, \bibinfo{pages}{5287--5297}.
\bibitem[{Milford and Wyeth(2012)}]{SeqSLAM}
\bibinfo{author}{Milford, M.J.}, \bibinfo{author}{Wyeth, G.F.},
  \bibinfo{year}{2012}.
\newblock \bibinfo{title}{Seqslam: Visual route-based navigation for sunny
  summer days and stormy winter nights}, in: \bibinfo{booktitle}{2012 IEEE
  international conference on robotics and automation},
  \bibinfo{organization}{IEEE}. pp. \bibinfo{pages}{1643--1649}.
\bibitem[{Pan et~al.(2021)Pan, Ji and Zhao}]{Pan_Tightly_coupled}
\bibinfo{author}{Pan, L.}, \bibinfo{author}{Ji, K.}, \bibinfo{author}{Zhao,
  J.}, \bibinfo{year}{2021}.
\newblock \bibinfo{title}{Tightly-coupled multi-sensor fusion for localization
  with lidar feature maps}, in: \bibinfo{booktitle}{2021 IEEE International
  Conference on Robotics and Automation (ICRA)}, \bibinfo{organization}{IEEE}.
  pp. \bibinfo{pages}{5215--5221}.
\bibitem[{Parlett(1998)}]{parlett1998symmetric}
\bibinfo{author}{Parlett, B.N.}, \bibinfo{year}{1998}.
\newblock \bibinfo{title}{The symmetric eigenvalue problem}.
\newblock \bibinfo{publisher}{SIAM}.
\bibitem[{Ramezani et~al.(2023)Ramezani, Wang, Knights, Li, Pounds and
  Moghadam}]{P_GAT}
\bibinfo{author}{Ramezani, M.}, \bibinfo{author}{Wang, L.},
  \bibinfo{author}{Knights, J.}, \bibinfo{author}{Li, Z.},
  \bibinfo{author}{Pounds, P.}, \bibinfo{author}{Moghadam, P.},
  \bibinfo{year}{2023}.
\newblock \bibinfo{title}{Pose-graph attentional graph neural network for lidar
  place recognition}.
\newblock \bibinfo{journal}{IEEE Robotics Automation Letters} .
\bibitem[{Sarlin et~al.(2020)Sarlin, DeTone, Malisiewicz and
  Rabinovich}]{Superglue}
\bibinfo{author}{Sarlin, P.E.}, \bibinfo{author}{DeTone, D.},
  \bibinfo{author}{Malisiewicz, T.}, \bibinfo{author}{Rabinovich, A.},
  \bibinfo{year}{2020}.
\newblock \bibinfo{title}{Superglue: Learning feature matching with graph
  neural networks}, in: \bibinfo{booktitle}{Proceedings of the IEEE/CVF
  conference on computer vision and pattern recognition}, pp.
  \bibinfo{pages}{4938--4947}.
\bibitem[{Serna and Marcotegui(2013)}]{serna2013urban_accessibility}
\bibinfo{author}{Serna, A.}, \bibinfo{author}{Marcotegui, B.},
  \bibinfo{year}{2013}.
\newblock \bibinfo{title}{Urban accessibility diagnosis from mobile laser
  scanning data}.
\newblock \bibinfo{journal}{ISPRS Journal of Photogrammetry and Remote Sensing}
  \bibinfo{volume}{84}, \bibinfo{pages}{23--32}.
\bibitem[{bin Shamsudin et~al.(2017)bin Shamsudin, Mizuno, Fujita, Ohno,
  Hamada, Westfechtel, Tadokoro and Amano}]{bin2017evaluation}
\bibinfo{author}{bin Shamsudin, A.U.}, \bibinfo{author}{Mizuno, N.},
  \bibinfo{author}{Fujita, J.}, \bibinfo{author}{Ohno, K.},
  \bibinfo{author}{Hamada, R.}, \bibinfo{author}{Westfechtel, T.},
  \bibinfo{author}{Tadokoro, S.}, \bibinfo{author}{Amano, H.},
  \bibinfo{year}{2017}.
\newblock \bibinfo{title}{Evaluation of lidar and gps based slam on fire
  disaster in petrochemical complexes}, in: \bibinfo{booktitle}{2017 IEEE
  International Symposium on Safety, Security and Rescue Robotics (SSRR)},
  \bibinfo{organization}{IEEE}. pp. \bibinfo{pages}{48--54}.
\bibitem[{Sharif(2021)}]{Sharif_Laser_privacy}
\bibinfo{author}{Sharif, M.H.}, \bibinfo{year}{2021}.
\newblock \bibinfo{title}{Laser-based algorithms meeting privacy in
  surveillance: A survey}.
\newblock \bibinfo{journal}{IEEE Access} \bibinfo{volume}{9},
  \bibinfo{pages}{92394--92419}.
\bibitem[{Sun et~al.(2020)Sun, Adolfsson, Magnusson, Andreasson, Posner and
  Duckett}]{Localising_Faster}
\bibinfo{author}{Sun, L.}, \bibinfo{author}{Adolfsson, D.},
  \bibinfo{author}{Magnusson, M.}, \bibinfo{author}{Andreasson, H.},
  \bibinfo{author}{Posner, I.}, \bibinfo{author}{Duckett, T.},
  \bibinfo{year}{2020}.
\newblock \bibinfo{title}{Localising faster: Efficient and precise lidar-based
  robot localisation in large-scale environments}, in: \bibinfo{booktitle}{2020
  IEEE international conference on robotics and automation (ICRA)},
  \bibinfo{organization}{IEEE}. pp. \bibinfo{pages}{4386--4392}.
\bibitem[{Tao et~al.(2022)Tao, Hu, Zhou, Xiao and Zhang}]{Seqpolar}
\bibinfo{author}{Tao, Q.}, \bibinfo{author}{Hu, Z.}, \bibinfo{author}{Zhou,
  Z.}, \bibinfo{author}{Xiao, H.}, \bibinfo{author}{Zhang, J.},
  \bibinfo{year}{2022}.
\newblock \bibinfo{title}{Seqpolar: sequence matching of polarized lidar map
  with hmm for intelligent vehicle localization}.
\newblock \bibinfo{journal}{IEEE Transactions on Vehicular Technology}
  \bibinfo{volume}{71}, \bibinfo{pages}{7071--7083}.
\bibitem[{Thrun(2002)}]{thrun2002probabilistic}
\bibinfo{author}{Thrun, S.}, \bibinfo{year}{2002}.
\newblock \bibinfo{title}{Probabilistic robotics}.
\newblock \bibinfo{journal}{Communications of the ACM} \bibinfo{volume}{45},
  \bibinfo{pages}{52--57}.
\bibitem[{Vaswani et~al.(2017)Vaswani, Shazeer, Parmar, Uszkoreit, Jones,
  Gomez, Kaiser and Polosukhin}]{Transformer}
\bibinfo{author}{Vaswani, A.}, \bibinfo{author}{Shazeer, N.},
  \bibinfo{author}{Parmar, N.}, \bibinfo{author}{Uszkoreit, J.},
  \bibinfo{author}{Jones, L.}, \bibinfo{author}{Gomez, A.N.},
  \bibinfo{author}{Kaiser, L.}, \bibinfo{author}{Polosukhin, I.},
  \bibinfo{year}{2017}.
\newblock \bibinfo{title}{Attention is all you need}.
\newblock \bibinfo{journal}{Advances in neural information processing systems}
  \bibinfo{volume}{30}.
\bibitem[{Vidanapathirana et~al.(2023)Vidanapathirana, Moghadam, Sridharan and
  Fookes}]{SGV}
\bibinfo{author}{Vidanapathirana, K.}, \bibinfo{author}{Moghadam, P.},
  \bibinfo{author}{Sridharan, S.}, \bibinfo{author}{Fookes, C.},
  \bibinfo{year}{2023}.
\newblock \bibinfo{title}{Spectral geometric verification: Re-ranking point
  cloud retrieval for metric localization}.
\newblock \bibinfo{journal}{IEEE Robotics Automation Letters} .
\bibitem[{Wan et~al.(2021)Wan, Zhang, He, Cui, Dai, Zhou and
  Huang}]{wan2021enhance}
\bibinfo{author}{Wan, Z.}, \bibinfo{author}{Zhang, Y.}, \bibinfo{author}{He,
  B.}, \bibinfo{author}{Cui, Z.}, \bibinfo{author}{Dai, W.},
  \bibinfo{author}{Zhou, L.}, \bibinfo{author}{Huang, G.},
  \bibinfo{year}{2021}.
\newblock \bibinfo{title}{Enhance accuracy: Sensitivity and uncertainty theory
  in lidar odometry and mapping}.
\newblock \bibinfo{journal}{arXiv preprint arXiv:2111.07723} .
\bibitem[{Wang et~al.(2021)Wang, Lou, Song, Yu and Tu}]{GM-Livox}
\bibinfo{author}{Wang, Y.}, \bibinfo{author}{Lou, Y.}, \bibinfo{author}{Song,
  W.}, \bibinfo{author}{Yu, H.}, \bibinfo{author}{Tu, Z.},
  \bibinfo{year}{2021}.
\newblock \bibinfo{title}{Gm-livox: An integrated framework for large-scale map
  construction with multiple non-repetitive scanning lidars} .
\bibitem[{Wu et~al.(2023)Wu, Zhang, Ke, Süsstrunk and
  Salzmann}]{Wu_Spatiotemporal_Self_supervised}
\bibinfo{author}{Wu, Y.}, \bibinfo{author}{Zhang, T.}, \bibinfo{author}{Ke,
  W.}, \bibinfo{author}{Süsstrunk, S.}, \bibinfo{author}{Salzmann, M.},
  \bibinfo{year}{2023}.
\newblock \bibinfo{title}{Spatiotemporal self-supervised learning for point
  clouds in the wild}.
\newblock \bibinfo{journal}{arXiv preprint arXiv:.16235} .
\bibitem[{Xie et~al.(2020)Xie, Gu, Guo, Qi, Guibas and Litany}]{Pointcontrast}
\bibinfo{author}{Xie, S.}, \bibinfo{author}{Gu, J.}, \bibinfo{author}{Guo, D.},
  \bibinfo{author}{Qi, C.R.}, \bibinfo{author}{Guibas, L.},
  \bibinfo{author}{Litany, O.}, \bibinfo{year}{2020}.
\newblock \bibinfo{title}{Pointcontrast: Unsupervised pre-training for 3d point
  cloud understanding}, in: \bibinfo{booktitle}{Computer Vision--ECCV 2020:
  16th European Conference, Glasgow, UK, August 23--28, 2020, Proceedings, Part
  III 16}, \bibinfo{organization}{Springer}. pp. \bibinfo{pages}{574--591}.
\bibitem[{Yang et~al.(2020)Yang, Shi and Carlone}]{Teaser}
\bibinfo{author}{Yang, H.}, \bibinfo{author}{Shi, J.},
  \bibinfo{author}{Carlone, L.}, \bibinfo{year}{2020}.
\newblock \bibinfo{title}{Teaser: Fast and certifiable point cloud
  registration}.
\newblock \bibinfo{journal}{IEEE Transactions on Robotics}
  \bibinfo{volume}{37}, \bibinfo{pages}{314--333}.
\bibitem[{Ye et~al.(2019)Ye, Chen and Liu}]{Ye_Tightly_coupled}
\bibinfo{author}{Ye, H.}, \bibinfo{author}{Chen, Y.}, \bibinfo{author}{Liu,
  M.}, \bibinfo{year}{2019}.
\newblock \bibinfo{title}{Tightly coupled 3d lidar inertial odometry and
  mapping}, in: \bibinfo{booktitle}{2019 International Conference on Robotics
  and Automation (ICRA)}, \bibinfo{organization}{IEEE}. pp.
  \bibinfo{pages}{3144--3150}.
\bibitem[{Yin et~al.(2019a)Yin, Tang, Ding, Wang and
  Xiong}]{Yin_A_failure_detection}
\bibinfo{author}{Yin, H.}, \bibinfo{author}{Tang, L.}, \bibinfo{author}{Ding,
  X.}, \bibinfo{author}{Wang, Y.}, \bibinfo{author}{Xiong, R.},
  \bibinfo{year}{2019}a.
\newblock \bibinfo{title}{A failure detection method for 3d lidar based
  localization}, in: \bibinfo{booktitle}{2019 Chinese Automation Congress
  (CAC)}, \bibinfo{organization}{IEEE}. pp. \bibinfo{pages}{4559--4563}.
\bibitem[{Yin et~al.(2019b)Yin, Wang, Ding, Tang, Huang and Xiong}]{LocNet}
\bibinfo{author}{Yin, H.}, \bibinfo{author}{Wang, Y.}, \bibinfo{author}{Ding,
  X.}, \bibinfo{author}{Tang, L.}, \bibinfo{author}{Huang, S.},
  \bibinfo{author}{Xiong, R.}, \bibinfo{year}{2019}b.
\newblock \bibinfo{title}{3d lidar-based global localization using siamese
  neural network}.
\newblock \bibinfo{journal}{IEEE Transactions on Intelligent Transportation
  Systems} \bibinfo{volume}{21}, \bibinfo{pages}{1380--1392}.
\bibitem[{Yin et~al.(2023a)Yin, Xu, Lu, Chen, Xiong, Shen, Stachniss and
  Wang}]{Global_LiDAR_Loc_Survey}
\bibinfo{author}{Yin, H.}, \bibinfo{author}{Xu, X.}, \bibinfo{author}{Lu, S.},
  \bibinfo{author}{Chen, X.}, \bibinfo{author}{Xiong, R.},
  \bibinfo{author}{Shen, S.}, \bibinfo{author}{Stachniss, C.},
  \bibinfo{author}{Wang, Y.}, \bibinfo{year}{2023}a.
\newblock \bibinfo{title}{A survey on global lidar localization: Challenges,
  advances and open problems}.
\newblock \bibinfo{journal}{arXiv preprint arXiv:.07433} .
\bibitem[{Yin et~al.(2023b)Yin, Abuduweili, Zhao, Xu, Liu and
  Scherer}]{BioSLAM}
\bibinfo{author}{Yin, P.}, \bibinfo{author}{Abuduweili, A.},
  \bibinfo{author}{Zhao, S.}, \bibinfo{author}{Xu, L.}, \bibinfo{author}{Liu,
  C.}, \bibinfo{author}{Scherer, S.}, \bibinfo{year}{2023}b.
\newblock \bibinfo{title}{Bioslam: A bioinspired lifelong memory system for
  general place recognition}.
\newblock \bibinfo{journal}{IEEE Transactions on Robotics} .
\bibitem[{Yin et~al.(2021)Yin, Wang, Egorov, Hou, Jia and
  Han}]{Seqspherevlad_v2}
\bibinfo{author}{Yin, P.}, \bibinfo{author}{Wang, F.}, \bibinfo{author}{Egorov,
  A.}, \bibinfo{author}{Hou, J.}, \bibinfo{author}{Jia, Z.},
  \bibinfo{author}{Han, J.}, \bibinfo{year}{2021}.
\newblock \bibinfo{title}{Fast sequence-matching enhanced viewpoint-invariant
  3-d place recognition}.
\newblock \bibinfo{journal}{IEEE Transactions on Industrial Electronics}
  \bibinfo{volume}{69}, \bibinfo{pages}{2127--2135}.
\bibitem[{Yin et~al.()Yin, Wang, Egorov, Hou, Zhang and
  Choset}]{Seqspherevlad_v1}
\bibinfo{author}{Yin, P.}, \bibinfo{author}{Wang, F.}, \bibinfo{author}{Egorov,
  A.}, \bibinfo{author}{Hou, J.}, \bibinfo{author}{Zhang, J.},
  \bibinfo{author}{Choset, H.}, .
\newblock \bibinfo{title}{Seqspherevlad: Sequence matching enhanced
  orientation-invariant place recognition}, in: \bibinfo{booktitle}{2020
  IEEE/RSJ International Conference on Intelligent Robots and Systems (IROS)},
  \bibinfo{publisher}{IEEE}. pp. \bibinfo{pages}{5024--5029}.
\bibitem[{Yuan et~al.(2017)Yuan, Zhang, Ding and Dong}]{yuan2017cooperative}
\bibinfo{author}{Yuan, J.}, \bibinfo{author}{Zhang, J.}, \bibinfo{author}{Ding,
  S.}, \bibinfo{author}{Dong, X.}, \bibinfo{year}{2017}.
\newblock \bibinfo{title}{Cooperative localization for disconnected sensor
  networks and a mobile robot in friendly environments}.
\newblock \bibinfo{journal}{Information Fusion} \bibinfo{volume}{37},
  \bibinfo{pages}{22--36}.
\bibitem[{Zhang et~al.(2012)Zhang, Zapata and Lepinay}]{Self_adaptive_MCL}
\bibinfo{author}{Zhang, L.}, \bibinfo{author}{Zapata, R.},
  \bibinfo{author}{Lepinay, P.}, \bibinfo{year}{2012}.
\newblock \bibinfo{title}{Self-adaptive monte carlo localization for mobile
  robots using range finders}.
\newblock \bibinfo{journal}{Robotica} \bibinfo{volume}{30},
  \bibinfo{pages}{229--244}.
\bibitem[{Zou et~al.(2023)Zou, Li, Wang, Liang, Wu, Wang, Yang and
  Dong}]{PatchAugNet}
\bibinfo{author}{Zou, X.}, \bibinfo{author}{Li, J.}, \bibinfo{author}{Wang,
  Y.}, \bibinfo{author}{Liang, F.}, \bibinfo{author}{Wu, W.},
  \bibinfo{author}{Wang, H.}, \bibinfo{author}{Yang, B.},
  \bibinfo{author}{Dong, Z.}, \bibinfo{year}{2023}.
\newblock \bibinfo{title}{Patchaugnet: Patch feature augmentation-based
  heterogeneous point cloud place recognition in large-scale street scenes}.
\newblock \bibinfo{journal}{ISPRS Journal of Photogrammetry Remote Sensing}
  \bibinfo{volume}{206}, \bibinfo{pages}{273--292}.

\end{thebibliography}
\end{document}